\PassOptionsToPackage{table}{xcolor}
\documentclass{article}
\usepackage{iclr2027_conference,times}

\usepackage[utf8]{inputenc}
\usepackage[T1]{fontenc}
\usepackage{CJKutf8}
\usepackage{microtype}
\usepackage{hyperref}
\usepackage{url}
\usepackage{booktabs}
\usepackage{amsfonts}
\usepackage{bbm}
\usepackage{nicefrac}
\usepackage{graphicx}
\usepackage{doi}
\usepackage{enumitem}
\usepackage{relsize}
\usepackage{amsmath,amssymb}
\usepackage{pgfplots}
\usepackage{tcolorbox}
\usepackage{subcaption}
\usepackage{pifont}

\usepackage{pdfpages}
\usepackage{multirow}
\usepackage[table]{xcolor}
\usepackage{array}
\usepackage{wrapfig}
\usepackage{needspace}
\usepackage{lineno}

\definecolor{darkblue}{rgb}{0,0,0.5}
\definecolor{mygreen}{HTML}{02790A}
\definecolor{myred}{HTML}{CF0000}

\definecolor{deltacol}{HTML}{02790A}
\newcolumntype{C}[1]{>{\centering\arraybackslash}p{#1}}

\title{Inject a Turn, Not a Token: Learning Where to Redirect LLM Reasoning for Self-Correction}

\author{Jiwon Jeon\thanks{Equal contribution.}, Myungsik Cho\footnotemark[1],
Jeonghye Kim, Youngchul Sung\\
Korea Advanced Institute of Science and Technology (KAIST) \\
\texttt{\{jiwon.jeon,ms.cho,jeonghye.kim,ycsung\}@kaist.ac.kr}
}

\hypersetup{
  colorlinks=true,
  citecolor=darkblue,
  linkcolor=darkblue,
  urlcolor=darkblue,
  pdftitle={Inject a Turn, Not a Token: Learning Where to Redirect LLM Reasoning for Self-Correction}
}

\iclrfinalcopy 
\begin{document}
\maketitle

\begin{abstract}
%
Large language models struggle to reliably identify and correct errors in their own reasoning. Unlike prior training-based approaches that fine-tune an LLM to generate revisions or feedback, we learn \emph{where} to intervene in failed rollouts and determine \emph{how} to redirect them. To choose where to intervene, we train a 130K-parameter MLP head on a frozen encoder using only final-answer correctness. For redirection, we find that under a fixed generation budget, the tested inline injections fail to reliably redirect reasoning, whereas presenting the same instruction in a new user turn changes the model's course. The resulting correction framework, UPTURN (User-Prompt TURN injection at a learned position), requires no LLM fine-tuning: it retains a selected prefix, injects an instruction as a new user turn, and regenerates the suffix. Across four reasoners and six math benchmarks, UPTURN improves average accuracy in every setting over a 32-rollout baseline. It outperforms self-correction and reflection-token injection baselines under matched correction budgets, transfers across reasoners and to GPQA-Diamond without retraining, and complements GRPO and self-correction methods.
\end{abstract}

\section{Introduction}
\label{sec:intro}
\begin{figure}[h]
    \centering
    \begingroup
    \setlength{\fboxsep}{4.75bp}
    \setlength{\fboxrule}{0.25bp}
    \resizebox{\linewidth}{!}{%
        \raisebox{\depth}{\fbox{%
            \begin{minipage}[c][242.5bp][c]{485bp}
                \centering
                \includegraphics[width=\linewidth,height=242.5bp,keepaspectratio]{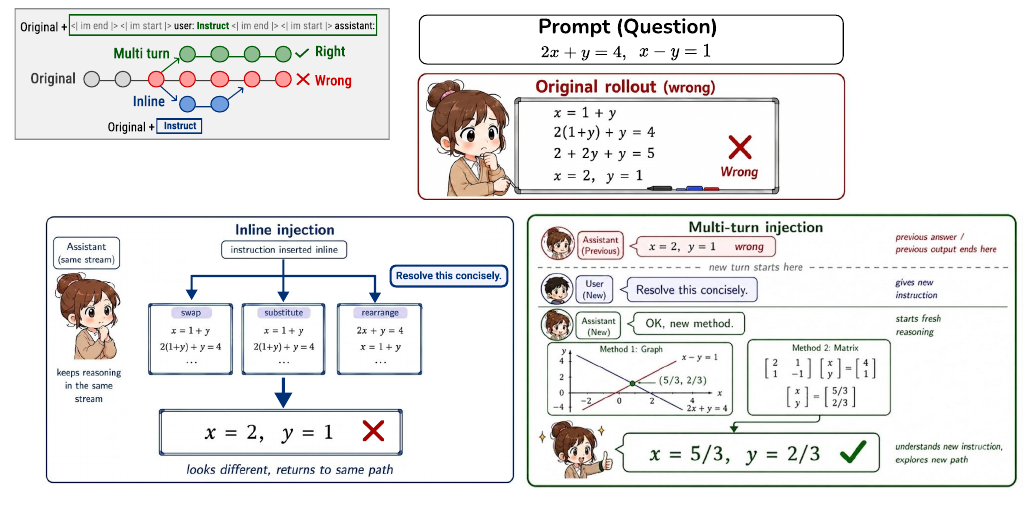}
            \end{minipage}%
        }}%
        \hspace{10bp}%
        \raisebox{\depth}{\fbox{%
            \begin{minipage}[c][242.5bp][c]{318bp}
                \centering
                \includegraphics[width=\linewidth,height=229bp,keepaspectratio]{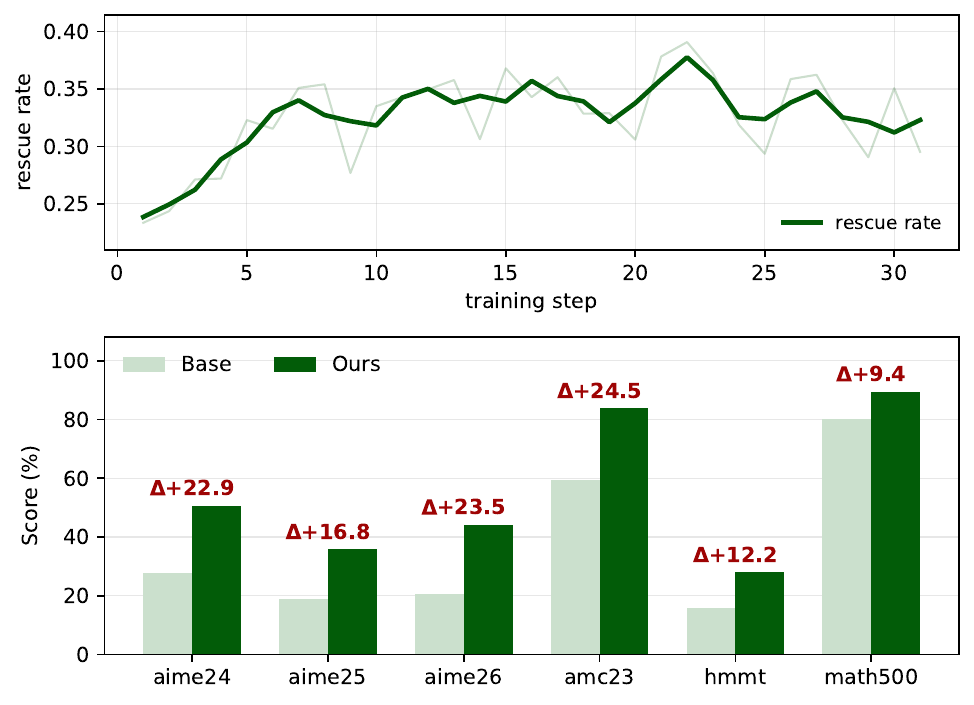}
            \end{minipage}%
        }}%
    }
    \endgroup
    \vspace{-1.6em}
    \caption{\textbf{Structure, not content alone, enables redirection.} \textbf{(Left)} The same instruction is absorbed inline but redirects reasoning in a new user turn. \textbf{(Right)} Learning where to intervene raises rescue rate (the fraction of failed rollouts corrected; top) and six-benchmark mean accuracy (bottom).}
\label{fig:intro}
\end{figure}

Large language models (LLMs) reason {\em autoregressively}, so a single mistake can derail subsequent reasoning, making effective correction important. Yet, given only its own rollout, a model may struggle to determine whether it has erred, and intrinsic self-correction without external feedback can even degrade reasoning accuracy~\citep{cannot_correct}. This motivates learning a correction signal from data. Existing methods do so by fine-tuning the reasoner to produce a revised answer~\citep{score, rise} or by training an external critic or verifier to provide step-level feedback or scores~\citep{lets, refiner}. Although effective, these methods require substantial correction data and typically optimize an LLM-scale component.


This raises a question: "Does correction really require training an entire LLM?" Repeated sampling shows that LLMs can solve far more problems than in a single attempt, with pass@$k$ increasing as $k$ grows~\citep{large_language_monkeys}. Yet individual rollouts can still fail: because autoregressive generation conditions each step on the preceding reasoning, even a single mistake can steer the subsequent trajectory away from a correct solution. A failed rollout need not involve many independent errors. Correction may therefore involve redirecting the model toward a solution it can already reach, rather than teaching it to reason {\em de novo}. We thus frame correction around two questions: \emph{where} to intervene in a failed rollout and \emph{how} to redirect it.


We study these questions in a \emph{fixed-budget prefix-regeneration} setting: after selecting a cut, we retain the prefix and regenerate only the suffix under the original response-token limit. Answering the \emph{where} question requires selecting a single cut position in a failed rollout rather than generating a revised answer or written feedback. We train a 130K-parameter MLP head on a frozen Qwen3-0.6B encoder to favor cut positions that more often lead to correct answers upon regeneration. To address \emph{how} to redirect the rollout, we consider token injection. Reflection tokens such as \texttt{``wait''} mark high-entropy forks where a reasoning path can branch in new directions~\citep{rule8020}, suggesting that they might steer a rollout. Indeed, s1~\citep{s1} improves accuracy by appending \texttt{``wait''} when the model attempts to stop, extending its reasoning. This suggests a simple recipe: learn where to intervene, inject a reflection token, and let the model continue.


However, our experiments reveal a critical limitation of inline injection under the fixed-budget protocol. Across the tested reflection tokens (\texttt{``wait''}, \texttt{``hmm''}), explicit instructions (\texttt{``Solve with a different way.''}), and even a task-irrelevant word (\texttt{``Turtle''}), inline injection fails to reliably redirect the rollout. The key difference is conversational \emph{structure}: the same instruction can redirect reasoning when delivered as a new user turn (Figure~\ref{fig:intro}, left), whereas inline delivery does not reliably do so. These results suggest that the model responds more effectively to an instruction delivered in a user turn than to the same text inserted into its own continuation. We refer to this structural transition from an assistant continuation to a new user turn as a \emph{turn boundary}.


Putting these two pieces together, we propose UPTURN (User-Prompt TURN injection at a learned position), a correction framework that requires no LLM fine-tuning. UPTURN keeps the reasoner and the 0.6B encoder frozen and trains a position policy, implemented as a 130K-parameter MLP head, to select where a fixed instruction is injected through a turn boundary. A separately trained 66K-parameter gate identifies which rollouts to correct at inference. Rather than identify the error or generate a fix, the position policy chooses where to redirect the rollout, while the frozen reasoner performs the correction itself. Averaged across four reasoners and six math benchmarks, UPTURN raises accuracy from 37.1\% to 55.3\% (Figure~\ref{fig:intro}, right). It outperforms self-correction and reflection-token injection baselines under matched correction budgets, as well as heuristic cuts and inline injection. Our contributions are threefold:
\begin{itemize}[leftmargin=1.0em]
    \item We show that conversational structure matters for correction: the tested inline injections fail to reliably redirect reasoning, whereas the same instruction delivered through a turn boundary can.
    \item We learn cut positions with a 130K-parameter head on a frozen encoder, using only regenerated rollouts' final-answer correctness, without step-level annotations or reasoner fine-tuning.
    \item We demonstrate consistent accuracy gains across four reasoners and six math benchmarks, transfer across reasoners and to GPQA-Diamond without retraining, and complementary gains on top of GRPO and existing self-correction methods.
\end{itemize}


\section{Related Work}
\label{sec:related}

\paragraph{Prompting-based self-correction.}
Prompting-based methods improve outputs without updating model weights, typically by having the model critique or reflect on its own attempt and then revise it~\citep{selfrefine}. Reflexion~\citep{reflexion}, for example, generates a verbal reflection on the previous attempt and uses it to guide a fresh attempt. These methods rely on the reasoner to diagnose its own errors, which can be unreliable for reasoning~\citep{cannot_correct, correctbench}. UPTURN asks the reasoner for neither a critique nor a rewrite: a learned position policy selects where to intervene, and a fixed user-turn instruction redirects the existing rollout.

\paragraph{Training-based self-correction.}
Training-based methods learn corrective signals from data, some by fine-tuning the reasoner to revise its own answers~\citep{rise}; SCoRe~\citep{score}, for example, trains the model with RL to produce a revised second attempt conditioned on its first. Others train a separate model to localize errors, such as a critic whose step-level feedback guides the reasoner's revision~\citep{refiner} or a process reward model that scores each step~\citep{lets}. These approaches typically train an LLM-scale component using correction data, such as revision pairs or step-level annotations. UPTURN instead trains lightweight heads to select which rollouts to correct and \emph{where} to intervene, rather than generating revisions, critiques, or step-level scores, while keeping the reasoner frozen.

\paragraph{Reflection-token injection.}
Several methods improve inference-time performance by injecting or promoting reflection tokens in a frozen model's reasoning. They append a token such as \texttt{``wait''} when the model attempts to stop~\citep{s1, continuethinking}, or encourage reflection tokens during generation on a predetermined schedule through direct insertion or logit upweighting~\citep{alphaone, cyclicreflex}. These methods steer an ongoing generation from within the model's own response, whereas we study post-hoc correction of completed rollouts flagged as likely incorrect. In this setting, we find that how the text is delivered matters: under a matched generation budget, the tested inline injections fail to reliably redirect reasoning even at learned cut positions (\S\ref{subsec:ablation}), whereas the same instruction delivered as a separate user turn does.

\section{Preliminaries}
\label{sec:preliminaries}

\paragraph{Autoregressive generation over chat turns.}
An autoregressive language model $p_\theta$ draws each token from the next-token distribution conditioned on all preceding tokens. We represent a conversation as alternating single-message user and assistant turns, written $[\textsc{user}{:}\,q]$ and $[\textsc{asst}{:}\,o]$ and concatenated with $\Vert$. Thus, a question $q$ and a response $o$ form the two-turn conversation $[\textsc{user}{:}\,q]\,\Vert\,[\textsc{asst}{:}\,o]$. Before generation, the chat template tokenizes the conversation and wraps each turn in special tokens marking where the turn begins and ends (e.g., \texttt{<|im\_start|>user} $\ldots$ \texttt{<|im\_end|>} for Qwen~\citep{qwen3}). The response is then generated token by token as
\begin{equation}
  o_t \sim p_\theta\big(\cdot \mid [\textsc{user}{:}\,q]\,\Vert\,[\textsc{asst}{:}\,o_{<t}]\big),
  \qquad t=1,\dots,T,
  \label{eq:chat-gen}
\end{equation}
where $o_{<t}=(o_1,\dots,o_{t-1})$, $T$ is the response length, and the assistant turn remains open.

\paragraph{Mathematical reasoning with LLMs.}
In the standard math setup, the model answers each problem with a chain-of-thought rollout $o$~\citep{cot} in a single assistant turn via Eq.~(\ref{eq:chat-gen}), with no further turns. The answer is the content of the last \texttt{\textbackslash boxed\{\}} expression in the rollout, and the reward is its exact match with the ground truth $a^\star$,
\begin{equation}
  R(q,o) \;=\; \mathlarger{\mathbf{1}}\!\left[\texttt{answer}(o) = a^\star\right] \;\in\; \{0,1\}.
  \label{eq:reward}
\end{equation}

\paragraph{Group Relative Policy Optimization.}
GRPO~\citep{grpo} is a policy-gradient method widely used to post-train LLM reasoners. Unlike PPO~\citep{ppo}, it removes the learned value network: for each input, it samples $G$ responses and forms advantages by centering and scaling their rewards with the group mean and standard deviation. It then applies PPO's clipped surrogate objective, typically with a KL penalty to a reference policy. We adapt this group-relative construction to train the position policy, with the modifications described in Sec.~\ref{subsec:training}.

\section{Motivation}
\label{sec:motivation}

\begin{figure}[h]
    \vspace{-0.3em}
    \centering
    {\captionsetup[subfigure]{skip=0pt}
    \subcaptionbox{Prefix-dependent recovery\label{fig:motivation-correct-prefix}}[\linewidth]{%
        \makebox[\linewidth][c]{%
            \includegraphics[width=0.247\linewidth]{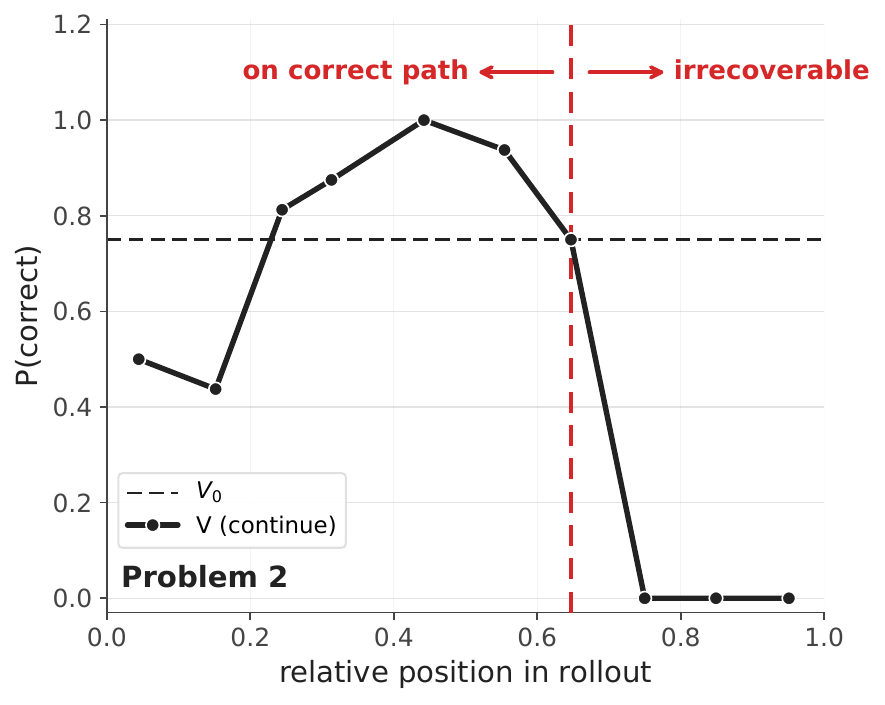}%
            \hspace{0.003\linewidth}%
            \includegraphics[width=0.247\linewidth]{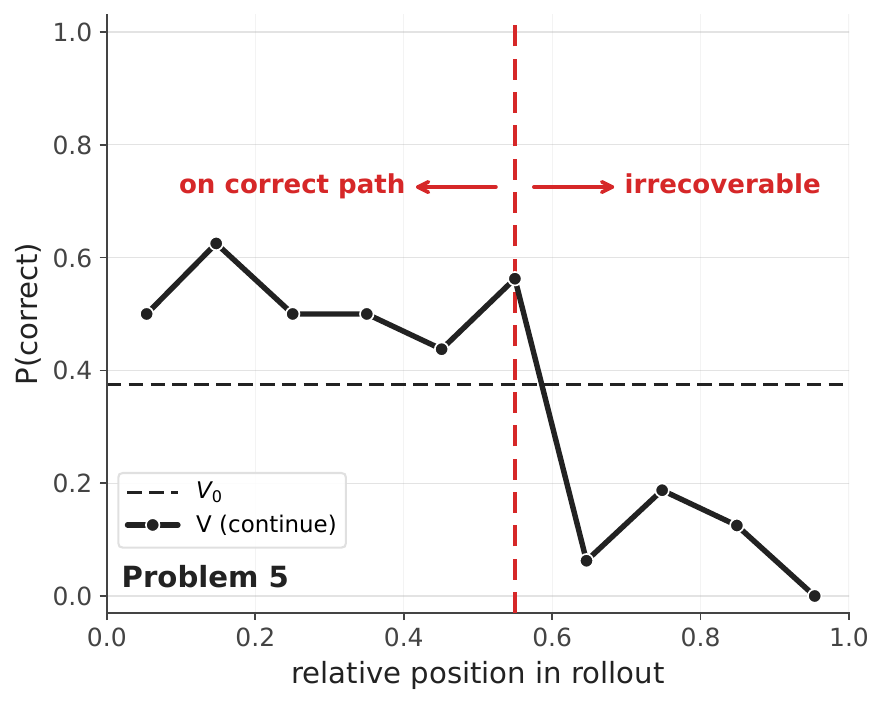}%
            \hspace{0.003\linewidth}%
            \includegraphics[width=0.247\linewidth]{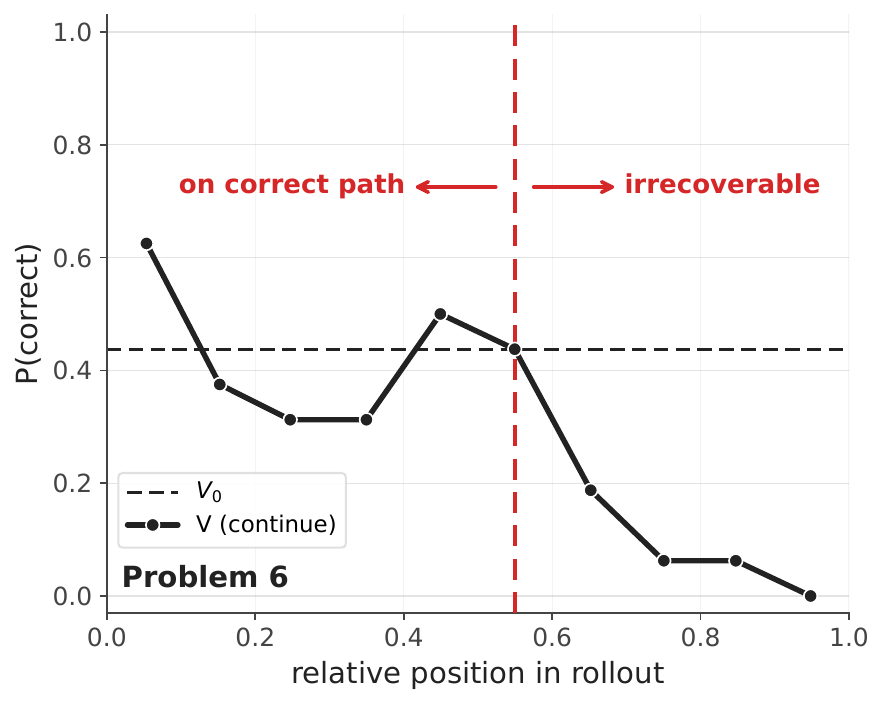}%
            \hspace{0.003\linewidth}%
            \includegraphics[width=0.247\linewidth]{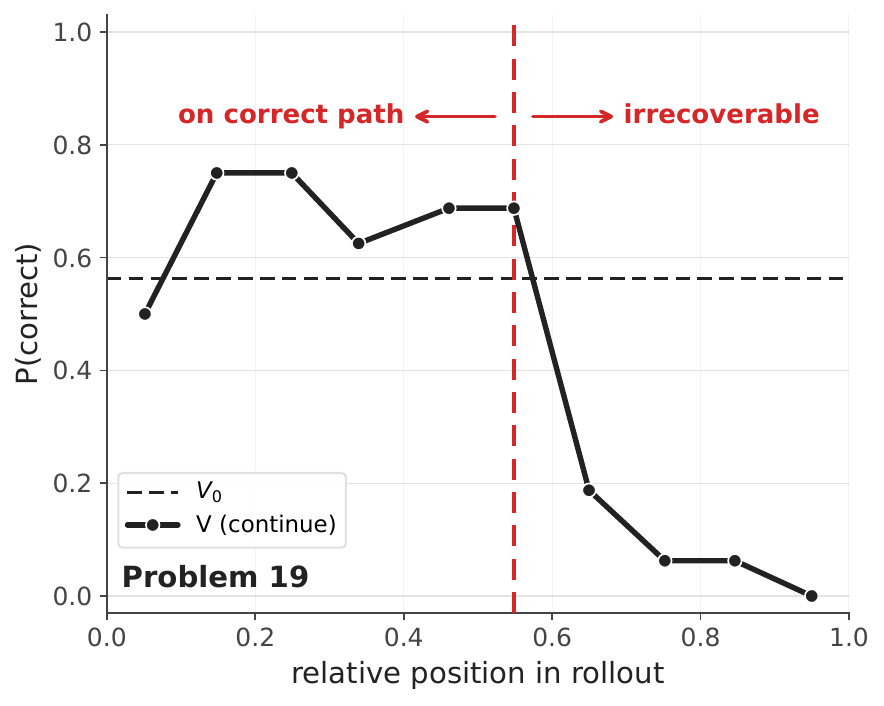}%
        }%
    }
    }
    \vspace{-1.1em}
    \begin{minipage}[t]{0.495\linewidth}
        \vspace{0pt}
        \centering
        {\captionsetup[subfigure]{skip=-3pt}
        \subcaptionbox{Position-wise rescue gap $(R-V)$\label{fig:motivation-r-minus-v}}[\linewidth]{%
            \includegraphics[width=\linewidth]{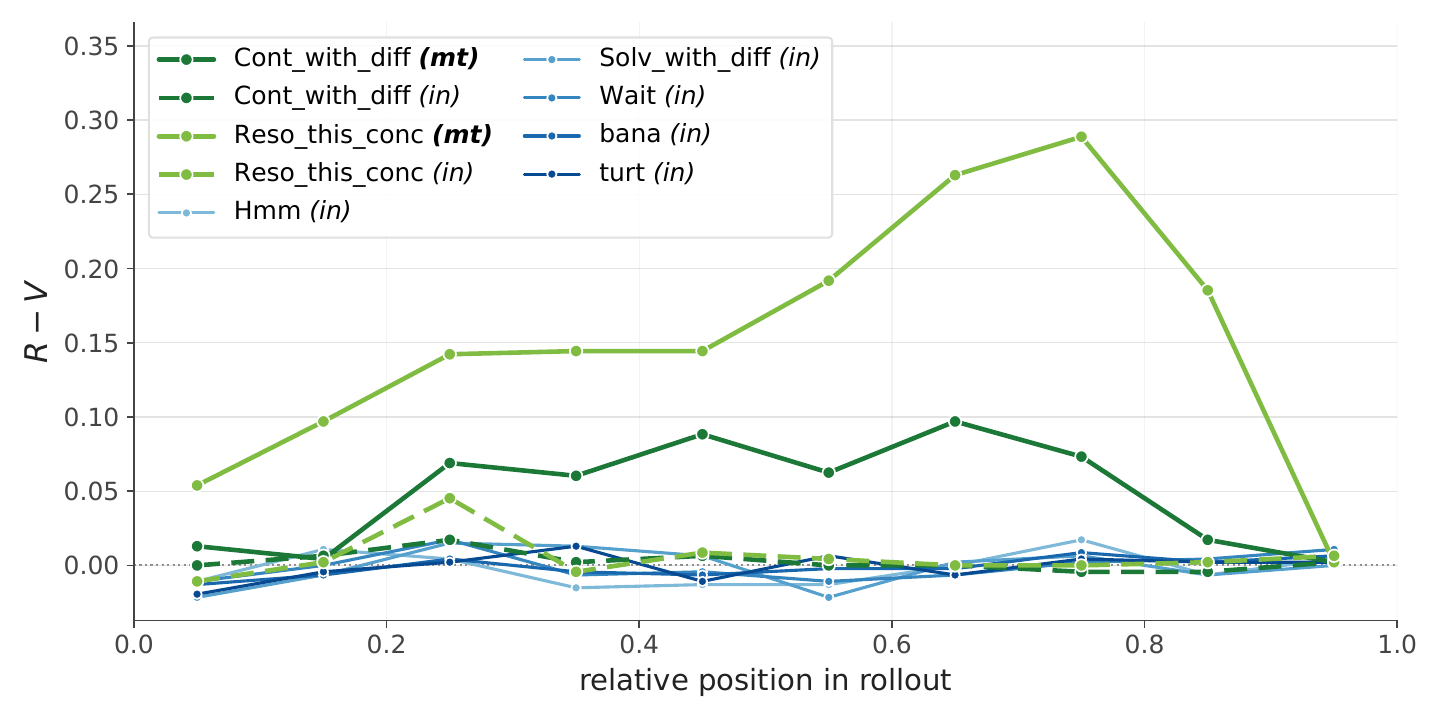}%
        }
        }
    \end{minipage}
    \hfill
    \begin{minipage}[t]{0.495\linewidth}
        \vspace{0pt}
        \centering
        {\captionsetup[subfigure]{skip=-3pt}
        \subcaptionbox{Oracle best rescue rate\label{fig:motivation-best-rescue}}[\linewidth]{%
            \includegraphics[width=\linewidth]{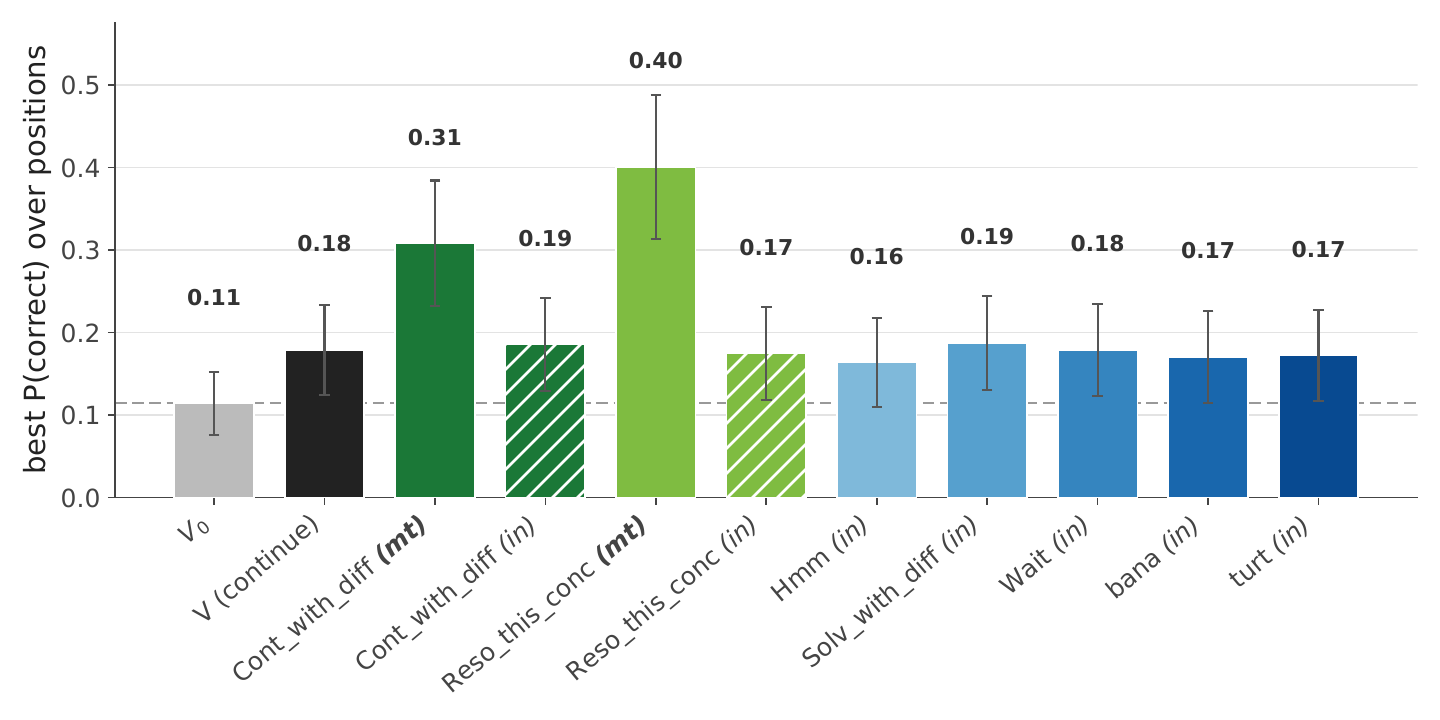}%
        }
        }
    \end{minipage}
    \caption{\textbf{Why turn-boundary injection is needed.}
    \textbf{(a)} Prefix value along failed rollouts, computed as the mean success rate over 16 continuations sampled from each prefix.
    \textbf{(b)} Position-wise rescue gap $R{-}V$ for inline (\textnormal{in}) and multi-turn (\textnormal{mt}) injection, where $R$ is post-injection success and $V$ is prefix value.
    \textbf{(c)} Oracle best rescue rate obtained by selecting, for each problem and phrase, the position with highest post-injection success.
    }
    \label{fig:motivation-evidence}
\end{figure}

\textit{Reinforcement Learning with Verifiable Rewards} (RLVR)~\citep{grpo, drgrpo} has enabled LLMs to produce high-quality mathematical solutions. Prior work~\citep{rule8020} shows that RLVR updates concentrate on high-entropy ``forking'' tokens (e.g., \texttt{Wait}, \texttt{Alternatively}), with smaller updates at the more numerous low-entropy tokens. These high-entropy tokens correspond to decision points where the reasoning can branch.

This observation motivates the following question for reasoning correction: If RLVR updates concentrate on such decision points during training, could intervening at analogous points during inference help correct a failed reasoning trace? A natural candidate is token injection: select a branching point and inject a reflection token to encourage a different continuation without retraining the reasoner.

\subsection{Recovery Depends on the Retained Prefix}
\label{subsec:correct_prefix_enable_recovery}

Failed traces may contain useful prefixes that can lead to correct answers. Restarting from scratch discards these prefixes, whereas regeneration from a suitable cut preserves them.


To test this, we collected failed Qwen3-4B (Thinking on) traces on AIME 2026 and extracted ten prefixes at equally spaced token positions from each trace. 
We sample 16 such continuations per prefix without any injected text and define the continuation success rate $V$ of each prefix as the fraction of continuations with correct final answers. As a reference, $V_0$ denotes the success rate when the model generates from the question alone, with no retained prefix. In the traces shown in Figure~\ref{fig:motivation-correct-prefix}, continuation success rates are high at earlier prefixes but drop sharply and remain low at later ones. This suggests that the choice of retained prefix can substantially affect the model's ability to recover.

The sharp decline may mark a branching point beyond which recovery becomes difficult. This motivates cutting before the transition to preserve a prefix from which the model can still recover.

\subsection{Inline vs.\ Multi-turn Injection}
\label{subsec:inline-vs-multiturn}

The previous subsection motivates \emph{where} to intervene: before a potential branching point marked by a sharp drop in continuation success. We now ask \emph{how} to redirect reasoning at a given cut position.

Using the recovery protocol in Sec.~\ref{subsec:correct_prefix_enable_recovery}, we compare injection variants at the sampled cut positions. At each position $t$, we retain the prefix $o_{\le t}$, discard the remaining suffix, and append the injected text $c$ within the same assistant turn before regeneration. We test reflection tokens such as \texttt{Wait} and \texttt{Hmm}, unrelated words such as \texttt{turtle} and \texttt{banana}, and longer instructions such as \texttt{``Solve with a different way.''}. 
For each rollout and cut position, only $c$ varies across the inline variants. Inline injection is expressed as
\begin{equation}
  \mathcal{C}_{\text{in}} = [\textsc{user}{:}\,q]\,\Vert\,[\textsc{asst}{:}\,o_{\le t},\,c],
\end{equation}
where $c=(c_1,\dots,c_m)$ denotes the injected text. Across the tested texts, inline injection yields similar continuation success rates. Surprisingly, these rates remain close to those obtained by regenerating from the prefix without injection (Figures~\ref{fig:motivation-r-minus-v} and~\ref{fig:motivation-best-rescue}).

We next deliver the same text $c$ through the chat template as a new user turn. We close the current assistant turn, insert a user turn containing $c$, and open a new assistant turn for regeneration:
\begin{equation}
  \mathcal{C}_{\text{mt}} = [\textsc{user}{:}\,q]\,\Vert\,[\textsc{asst}{:}\,o_{\le t}]\,\Vert\,[\textsc{user}{:}\,c]\,\Vert\,[\textsc{asst}{:}\,\cdot].
\end{equation}
This multi-turn format yields substantially higher continuation success rates than inline injection.

These results show that the injection format matters beyond instruction wording. Both contexts use the same frozen model $p_\theta$ and injected text $c$, but differ in their conditioning format. In $\mathcal{C}_{\text{in}}$, the short $c$ ($m\ll t$) follows a long sequence of assistant tokens and may therefore be interpreted as part of the ongoing reasoning rather than as a revision signal. In $\mathcal{C}_{\text{mt}}$, the chat template instead presents the same $c$ as an {\em external} instruction in a new user turn, which may help redirect the continuation. Figure~\ref{fig:motivation} in Appendix~\ref{app:motivation-example} illustrates this contrast: inline continuation stays on the original path, whereas a new user turn redirects reasoning.

\section{Method}
\label{sec:method}
\begin{figure}[h]
    \vspace{-.8em}
    \centering
    \includegraphics[width=0.95\linewidth]{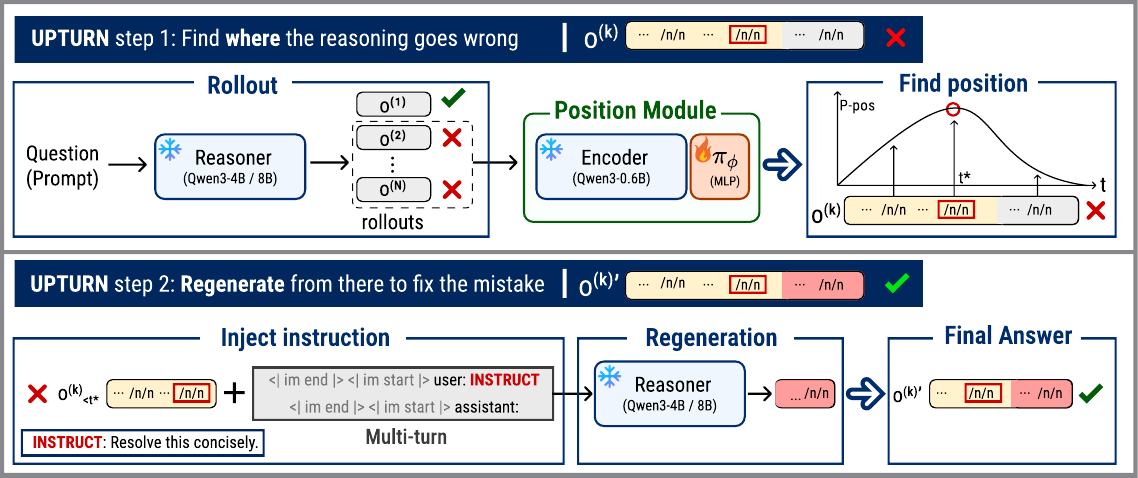}
    \vspace{-.5em}
    \caption{\textbf{Overview of UPTURN.}
    \textbf{(Step 1)} A frozen 0.6B encoder and MLP head $f_\phi$ score paragraph boundaries (\texttt{\textbackslash n\textbackslash n}) in failed rollouts ({\color{myred}\ding{55}}); the policy selects a cut $t^\star$.
    \textbf{(Step 2)} UPTURN retains the prefix, inserts a fixed instruction as a new user turn, and regenerates in a new assistant turn, recovering a correct answer ({\color{mygreen}\ding{51}}).
    }
    \label{fig:overview}
\end{figure}

Motivated by \S\ref{sec:motivation}, we correct a failed rollout by discarding the suffix after a selected position, inserting a fixed instruction $c$ as a new user turn, and letting the frozen reasoner regenerate (Figure~\ref{fig:overview}). The reasoner $p_\theta$ (e.g., Qwen3-4B) produces both the original and revised rollouts, while the position policy $\pi_\phi$ selects the cut. All LLM parameters and the injection instruction remain fixed; only lightweight heads are trained.

\subsection{Position policy}
\label{subsec:position_policy}

Given a question $q$ and a failed rollout $o$, a frozen Qwen3-0.6B causal encoder $E$ produces a hidden state $h_t$ at each token $o_t$. Because the encoder is causal, each candidate's score $f_\phi(h_t)$ depends only on the question and retained prefix $o_{\le t}$, not on the discarded suffix. To reduce the cost of evaluating regeneration rewards at every token, we restrict cuts to paragraph boundaries (\texttt{\textbackslash n\textbackslash n}), natural breaks between reasoning steps. Let $\mathcal{T}(o)$ denote the token positions at the ends of these boundaries. Appendix~\ref{app:entropy} shows that next-token entropy is elevated at paragraph boundaries, supporting these candidate cuts. Cutting at $t$ retains $o_{\le t}$ and discards $o_{>t}$. We train only a two-layer MLP head $f_\phi$ ($130$K parameters) and normalize its scores over candidates to obtain the policy:
\begin{equation}
  \pi_\phi(t \mid q, o)
  = \frac{\exp\!\big(f_\phi(h_{t})\big)}{\sum_{t'\in\mathcal{T}(o)}\exp\!\big(f_\phi(h_{t'})\big)}.
  \label{eq:policy}
\end{equation}

\subsection{Training the position policy}
\label{subsec:training}

\paragraph{Collecting failed rollouts.}
We sample $N$ rollouts per prompt and train on all failures ($R(q,o)=0$). Since all-success prompts offer nothing to correct, we sample additional prompt batches until obtaining $B$ prompts with at least one failure, similar to dynamic sampling~\citep{dapo}.

\paragraph{Multi-turn instruction injection.} 
Based on Sec.~\ref{subsec:inline-vs-multiturn}, given a failed rollout and a cut position $t$, we close the assistant turn after the prefix $o_{\le t}$, insert the fixed instruction $c$ (\texttt{``Resolve this concisely.''}) as a new user turn, and open a new assistant turn,
\begin{equation}
  \mathrm{ctx}(q, o_{\le t}, c)=
  [\textsc{user}{:}\,q]\,\Vert\,[\textsc{asst}{:}\,o_{\le t}]\,\Vert\,
  [\textsc{user}{:}\,c]\,\Vert\,[\textsc{asst}{:}\,\cdot],
  \label{eq:ctx}
\end{equation}
Here, $[\textsc{asst}{:}\,\cdot]$ denotes the new assistant turn from which generation begins. We use this injection format during both training and inference. The reasoner generates at most $L_{\max}{-}t$ suffix tokens, keeping the retained prefix and regenerated suffix within the original response-token budget $L_{\max}$.

\paragraph{Reward and objective.}
We reward cuts by regeneration success. For each failed rollout, we sample $G$ cuts $\{t^{(1)},\dots,t^{(G)}\}$ without replacement from $\pi_\phi(\cdot\mid q,o)$ as one GRPO group. At each $t^{(i)}$, the reasoner generates $M$ suffixes $\tilde{o}^{(i,s)}\sim p_\theta(\cdot\mid\mathrm{ctx}(q,o_{\le t^{(i)}},c))$. We score each cut by averaging the final-answer correctness of its $M$ regenerated suffixes:
\vspace{-0.1em}
\begin{equation}
  r_i=\frac{1}{M}\sum_{s=1}^{M} R\big(q,\tilde{o}^{(i,s)}\big).
  \label{eq:cut-reward}
  \vspace{-0.1em}
\end{equation}
With $p_\theta$ and $c$ fixed, averaging correctness over $M$ regenerated suffixes reduces reward-estimation variance. Unlike standard GRPO, which groups independent responses, we group cuts from one failed rollout to compare positions while holding the trajectory fixed. Following Dr.GRPO~\citep{drgrpo}, we drop standard-deviation normalization and use the mean-subtracted advantage $\hat{A}_i{=}r_i{-}\frac{1}{G}\sum_{j=1}^{G} r_j$, with importance ratio $\rho_i{=}\pi_\phi(t^{(i)}\mid q,o)/\pi_{\phi_{\text{old}}}(t^{(i)}\mid q,o)$. The objective is
\vspace{-0.1em}
\begin{equation}
  \mathcal{J}(\phi) = \mathbb{E}\!\left[\frac{1}{G}\sum_{i=1}^{G}
  \min\!\big(\rho_i\,\hat{A}_i,\ \operatorname{clip}(\rho_i, 1-\varepsilon, 1+\varepsilon)\,\hat{A}_i\big)
  \;+\; \beta_{\mathrm{ent}}\,\mathcal{H}\!\big[\pi_\phi(\cdot\mid q,o)\big]\right].
  \label{eq:objective}
  \vspace{-0.1em}
\end{equation}
We omit the reference-KL term and use an entropy bonus for cut exploration. Advantages computed within a rollout favor cuts with higher rescue rates than other cuts in that trace. 

\subsection{Gated Inference}
\label{subsec:inference}
At inference, ground-truth answers are unavailable, so we use a lightweight \emph{gate} $g_\psi$ to estimate whether each rollout is correct. Rollouts without a parseable \texttt{\textbackslash boxed\{\}} answer receive a gate score of zero. For the rest, the same frozen Qwen3-0.6B encoder $E$ processes the chat-templated question and rollout, and an MLP predicts correctness from the hidden state $h_{\mathrm{box}}$ at the final \texttt{\textbackslash boxed\{\}} token. With $\kappa$ denoting the calibration temperature, the gate is defined as
\begin{equation}
  g_\psi(q,o)=
  \begin{cases}
    0, & \text{if } o \text{ has no parseable answer},\\
    \sigma\!\big(\mathrm{MLP}_\psi(h_{\mathrm{box}})/\kappa\big), & \text{otherwise},
  \end{cases}
  \label{eq:gate}
\end{equation}
The gate is a 66K-parameter MLP trained once on 2,874 correctness-labeled rollouts and reused across all reasoners. Appendix~\ref{app:gate} provides training and calibration details, evaluates gate ranking quality, compares learned and ground-truth gating, and examines threshold sensitivity.

Rollouts with $g_\psi(q,o)\ge\gamma$ are retained unchanged. Those below the fixed threshold $\gamma{=}0.4$ are corrected once by sampling $t^\star\sim\pi_\phi(\cdot\mid q,o)$, injecting $c$ via Eq.~\eqref{eq:ctx}, and regenerating. Each regenerated rollout replaces its original, preserving the pool size.

\section{Experiments}
\label{sec:experiments}
We ask four questions: 
(1) Can UPTURN correct rollouts to improve accuracy across models and benchmarks within a fixed generation budget?
(2) How does it compare with self-correction and reflection-token injection? 
(3) Does it transfer across reasoners and domains and complement GRPO and self-correction? 
(4) How do learned cuts and turn-boundary injection affect performance?

\subsection{Setup}
\label{subsec:setup}

\paragraph{Benchmarks, metrics, and reasoners.}
We train each position policy for 30 steps on DAPO-17K, a 17K-prompt DAPO-Math subset~\citep{dapo}. Evaluation covers AIME24, AIME25, AIME26, AMC23, HMMT, and MATH500, all disjoint from training. \textbf{Avg} is the mean per-question accuracy of the final candidate pool and \textbf{Pass} the fraction of questions with at least one correct candidate. We evaluate four frozen reasoners in the main table: Qwen3-4B and Qwen3-8B with thinking enabled, DeepSeek-R1-0528-Qwen3-8B (DeepSeek-R1-Qwen3-8B), and QwQ-32B; two additional reasoners are reported in Appendix~\ref{app:other-models}. Unless stated otherwise, each reasoner uses a position policy trained on its own rollouts, while the 66K gate is reused unchanged across all models.

\paragraph{Decoding budgets.}
Base and UPTURN share the frozen reasoner and sampling parameters. \textbf{Base@32} draws 32 independent full rollouts. \textbf{UPTURN@16} draws 16 rollouts and corrects each gate-flagged rollout once from $t^\star{\sim}\pi_\phi(\cdot{\mid}q,o)$, adding up to 16 generations. Thus, both methods allow up to 32 generations per question. Base therefore receives a larger maximum generated-token allowance: each correction reuses $o_{\le t^\star}$ and generates at most $L_{\max}{-}t^\star$ new tokens instead of a full rollout. Results at larger $L_{\max}$ are in Appendix~\ref{app:lmax-scale}. Under gated inference (\S\ref{subsec:inference}), corrections replace their originals, leaving 16 rollouts for evaluation. Sampling settings appear in Appendix~\ref{app:impl}. We also measure end-to-end runtime, including gating and position-selection overhead (Appendix~\ref{app:throughput}).

\subsection{Main results}
\begin{table}[t!]
    \vspace{-0.3em}
    \centering
    \setlength{\tabcolsep}{2pt}
    \renewcommand{\arraystretch}{0.95}
    \caption{Results on six math benchmarks. \textbf{Base@32} uses 32 rollouts; \textbf{UPTURN@16} uses 16 rollouts and corrects each gate-flagged one once. $\Delta$ denotes UPTURN's gain over Base; \textbf{bold} marks the higher score.}
    \vspace{-0.8em}
    \resizebox{\textwidth}{!}{%
    \begin{tabular}{p{3.3cm}
        C{0.9cm}C{0.9cm} C{3pt}
        C{0.9cm}C{0.9cm} C{3pt}
        C{0.9cm}C{0.9cm} C{3pt}
        C{0.9cm}C{0.9cm} C{3pt}
        C{0.9cm}C{0.9cm} C{3pt}
        C{0.9cm}C{0.9cm}}
        \toprule
        \multirow{2}{*}{\textbf{Method}}
            & \multicolumn{2}{c}{\textbf{AIME24}} &
            & \multicolumn{2}{c}{\textbf{AIME25}} &
            & \multicolumn{2}{c}{\textbf{AIME26}} &
            & \multicolumn{2}{c}{\textbf{AMC23}} &
            & \multicolumn{2}{c}{\textbf{HMMT}} &
            & \multicolumn{2}{c}{\textbf{MATH500}} \\
        \cmidrule(lr){2-3} \cmidrule(lr){5-6} \cmidrule(lr){8-9} \cmidrule(lr){11-12} \cmidrule(lr){14-15} \cmidrule(lr){17-18}
             & {\scriptsize Avg} & {\scriptsize Pass} &
            & {\scriptsize Avg} & {\scriptsize Pass} &
            & {\scriptsize Avg} & {\scriptsize Pass} &
            & {\scriptsize Avg} & {\scriptsize Pass} &
            & {\scriptsize Avg} & {\scriptsize Pass} &
            & {\scriptsize Avg} & {\scriptsize Pass} \\
        \midrule

        \multicolumn{18}{@{}l}{\textit{Qwen3-4B}} \\
        \quad Base@32 & 32.6 & 63.3 & & 21.8 & 46.7 & & 26.7 & 43.3 & & 67.0 & 92.5 & & 19.2 & 27.3 & & 85.0 & 93.2 \\
        \quad \textbf{UPTURN@16} & \textbf{50.8} & \textbf{83.3} & & \textbf{39.0} & \textbf{56.7} & & \textbf{44.2} & \textbf{70.0} & & \textbf{84.2} & \textbf{95.0} & & \textbf{28.0} & \textbf{39.4} & & \textbf{91.0} & \textbf{95.8} \\
        \quad $\Delta$ & \cellcolor{gray!10}\textcolor{deltacol}{+18.2} & \cellcolor{gray!10}\textcolor{deltacol}{+20.0} & \cellcolor{gray!10} & \cellcolor{gray!10}\textcolor{deltacol}{+17.2} & \cellcolor{gray!10}\textcolor{deltacol}{+10.0} & \cellcolor{gray!10} & \cellcolor{gray!10}\textcolor{deltacol}{+17.5} & \cellcolor{gray!10}\textcolor{deltacol}{+26.7} & \cellcolor{gray!10} & \cellcolor{gray!10}\textcolor{deltacol}{+17.2} & \cellcolor{gray!10}\textcolor{deltacol}{+2.5} & \cellcolor{gray!10} & \cellcolor{gray!10}\textcolor{deltacol}{+8.8} & \cellcolor{gray!10}\textcolor{deltacol}{+12.1} & \cellcolor{gray!10} & \cellcolor{gray!10}\textcolor{deltacol}{+6.0} & \cellcolor{gray!10}\textcolor{deltacol}{+2.6} \\
        \addlinespace[3pt]
        \midrule

        \multicolumn{18}{@{}l}{\textit{Qwen3-8B}} \\
        \quad Base@32 & 35.8 & 56.7 & & 20.9 & 43.3 & & 24.7 & 50.0 & & 64.8 & 92.5 & & 19.0 & 27.3 & & 83.1 & 93.0 \\
        \quad \textbf{UPTURN@16} & \textbf{53.8} & \textbf{70.0} & & \textbf{39.2} & \textbf{56.7} & & \textbf{47.7} & \textbf{66.7} & & \textbf{85.3} & \textbf{95.0} & & \textbf{32.2} & \textbf{51.5} & & \textbf{92.0} & \textbf{96.4} \\
        \quad $\Delta$ & \cellcolor{gray!10}\textcolor{deltacol}{+17.9} & \cellcolor{gray!10}\textcolor{deltacol}{+13.3} & \cellcolor{gray!10} & \cellcolor{gray!10}\textcolor{deltacol}{+18.2} & \cellcolor{gray!10}\textcolor{deltacol}{+13.3} & \cellcolor{gray!10} & \cellcolor{gray!10}\textcolor{deltacol}{+23.0} & \cellcolor{gray!10}\textcolor{deltacol}{+16.7} & \cellcolor{gray!10} & \cellcolor{gray!10}\textcolor{deltacol}{+20.5} & \cellcolor{gray!10}\textcolor{deltacol}{+2.5} & \cellcolor{gray!10} & \cellcolor{gray!10}\textcolor{deltacol}{+13.2} & \cellcolor{gray!10}\textcolor{deltacol}{+24.2} & \cellcolor{gray!10} & \cellcolor{gray!10}\textcolor{deltacol}{+8.8} & \cellcolor{gray!10}\textcolor{deltacol}{+3.4} \\
        \addlinespace[3pt]
        \midrule

        \multicolumn{18}{@{}l}{\textit{DeepSeek-R1-Qwen3-8B}} \\
        \quad Base@32 & 5.8 & 16.7 & & 7.1 & 13.3 & & 3.5 & 6.7 & & 32.3 & 47.5 & & 2.0 & 9.1 & & 64.5 & 79.8 \\
        \quad \textbf{UPTURN@16} & \textbf{43.8} & \textbf{70.0} & & \textbf{31.2} & \textbf{53.3} & & \textbf{35.2} & \textbf{56.7} & & \textbf{79.5} & \textbf{97.5} & & \textbf{24.6} & \textbf{48.5} & & \textbf{82.8} & \textbf{95.6} \\
        \quad $\Delta$ & \cellcolor{gray!10}\textcolor{deltacol}{+37.9} & \cellcolor{gray!10}\textcolor{deltacol}{+53.3} & \cellcolor{gray!10} & \cellcolor{gray!10}\textcolor{deltacol}{+24.2} & \cellcolor{gray!10}\textcolor{deltacol}{+40.0} & \cellcolor{gray!10} & \cellcolor{gray!10}\textcolor{deltacol}{+31.7} & \cellcolor{gray!10}\textcolor{deltacol}{+50.0} & \cellcolor{gray!10} & \cellcolor{gray!10}\textcolor{deltacol}{+47.2} & \cellcolor{gray!10}\textcolor{deltacol}{+50.0} & \cellcolor{gray!10} & \cellcolor{gray!10}\textcolor{deltacol}{+22.6} & \cellcolor{gray!10}\textcolor{deltacol}{+39.4} & \cellcolor{gray!10} & \cellcolor{gray!10}\textcolor{deltacol}{+18.3} & \cellcolor{gray!10}\textcolor{deltacol}{+15.8} \\
        \addlinespace[3pt]
        \midrule

        \multicolumn{18}{@{}l}{\textit{QwQ-32B}} \\
        \quad Base@32 & 36.7 & 63.3 & & 26.2 & 46.7 & & 27.4 & 56.7 & & 73.0 & 87.5 & & 22.7 & 33.3 & & 87.8 & 95.0 \\
        \quad \textbf{UPTURN@16} & \textbf{54.4} & \textbf{70.0} & & \textbf{33.8} & \textbf{56.7} & & \textbf{49.4} & \textbf{63.3} & & \textbf{86.2} & \textbf{92.5} & & \textbf{27.1} & 33.3 & & \textbf{92.2} & \textbf{96.2} \\
        \quad $\Delta$ & \cellcolor{gray!10}\textcolor{deltacol}{+17.7} & \cellcolor{gray!10}\textcolor{deltacol}{+6.7} & \cellcolor{gray!10} & \cellcolor{gray!10}\textcolor{deltacol}{+7.5} & \cellcolor{gray!10}\textcolor{deltacol}{+10.0} & \cellcolor{gray!10} & \cellcolor{gray!10}\textcolor{deltacol}{+22.0} & \cellcolor{gray!10}\textcolor{deltacol}{+6.7} & \cellcolor{gray!10} & \cellcolor{gray!10}\textcolor{deltacol}{+13.2} & \cellcolor{gray!10}\textcolor{deltacol}{+5.0} & \cellcolor{gray!10} & \cellcolor{gray!10}\textcolor{deltacol}{+4.4} & \cellcolor{gray!10}{0.0} & \cellcolor{gray!10} & \cellcolor{gray!10}\textcolor{deltacol}{+4.3} & \cellcolor{gray!10}\textcolor{deltacol}{+1.2} \\
        \bottomrule
    \end{tabular}%
    }
    \label{tab:main_results}
\end{table}

\label{subsec:main_results}

Table~\ref{tab:main_results} compares \textbf{UPTURN@16} with \textbf{Base@32} across six benchmarks and four reasoners spanning model families and sizes from 4B to 32B. Despite Base receiving 32 full rollouts, UPTURN improves average accuracy on every benchmark and pass rate in all but one setting. This consistency supports the effectiveness of its correction mechanism. Gains are pronounced on AIME, whereas accuracy gains are smallest on MATH500 for every reasoner, where Base accuracy is already high.

The largest improvements occur on DeepSeek-R1-Qwen3-8B, reaching $+47.2$ points in average accuracy on AMC23 and $+53.3$ points in pass rate on AIME24. These gains highlight the effectiveness of redirecting existing rollouts under a constrained generation budget, where DeepSeek-R1-Qwen3-8B's long reasoning traces are frequently truncated.
The benefits also persist on QwQ-32B: average accuracy improves on all six benchmarks, while pass rate improves on five and remains unchanged on HMMT. Results on two additional RL-finetuned reasoners show the same trend (Appendix~\ref{app:other-models}). We also assess robustness across three seeds for policy training and rollout sampling (Appendix~\ref{app:seedrep}).

\subsection{Comparison with self-correction baselines}
\label{subsec:baselines}
Table~\ref{tab:budget_matched} compares UPTURN with self-correction methods Reflexion~\citep{reflexion} and SCoRe~\citep{score}, and reflection-token injection methods AlphaOne~\citep{alphaone} and CyclicReflex~\citep{cyclicreflex}, averaged over Qwen3-4B, Qwen3-8B, and DeepSeek-R1-Qwen3-8B (full results in Appendices~\ref{app:baseline-detail} and~\ref{app:reflect3}). All methods start from the same 16 rollouts and correct each gate-flagged one once, following the budget protocol in Sec.~\ref{subsec:setup}. SCoRe conditions on the full first attempt, leaving almost no room for revision at $L_{\max}$, so we also compare methods at $2L_{\max}$.

\ifdefined\UPTURNcolm
\begin{table*}[h]
    \vspace{-0.5em}
\else
\begin{table*}[t]
    \vspace{-0.3em}
\fi
    \centering
    \setlength{\tabcolsep}{8pt}
    \renewcommand{\arraystretch}{0.95}
    \caption{\textbf{Comparison with other baselines.} Avg\,/\,Pass averaged over six benchmarks and all reasoners except QwQ-32B, at two budgets ($L_{\max}$ and $2L_{\max}$). Parentheses give UPTURN's gain over the best baseline at each budget; \textbf{bold} marks the best scores. $^\dagger$SCoRe requires $2L_{\max}$ as it conditions on the full first attempt.}
    \label{tab:budget_matched}
    \vspace{-0.8em}
    \fontsize{8pt}{8pt}\selectfont
    \resizebox{\textwidth}{!}{%
    \begin{tabular}{@{}l c c c c c@{}}
        \toprule
        & \multicolumn{2}{c}{\textit{Self-correction}} & \multicolumn{2}{c}{\textit{Reflection-token injection}} & \\
        \cmidrule(lr){2-3}\cmidrule(lr){4-5}
        \textbf{Budget} & \textbf{Reflexion@16} & \textbf{SCoRe@16} & \textbf{AlphaOne@16} & \textbf{CyclicReflex@16} & \textbf{UPTURN@16} \\
        \midrule
        $L_{\max}$ & 43.6 / 61.4 & --$^\dagger$ & 32.6 / 45.2 & 35.7 / 49.7 & \textbf{59.8} / \textbf{77.2} (\textcolor{deltacol}{$+16.2$ / $+15.9$}) \\
        $2L_{\max}$ & 55.3 / 77.5 & 52.3 / 79.9 & 55.0 / 71.8 & 57.2 / 73.5 & \textbf{62.2} / \textbf{80.7} (\textcolor{deltacol}{$+4.9$ / $+0.8$}) \\
        \bottomrule
    \end{tabular}%
    }
    \vspace{-0.8em}
\end{table*}

UPTURN performs best at both budgets, with a larger advantage under the tighter token budget. At $L_{\max}$, it exceeds the strongest baseline, Reflexion, by $+16.2$ points in average accuracy and $+15.9$ in pass rate, while the two injection methods remain close to Base@32 (Table~\ref{tab:main_results}). Increasing the budget to $2L_{\max}$ improves the baselines evaluated at both budgets and narrows the gap, but UPTURN retains gains of $+4.9$ points in average accuracy and $+0.8$ in pass rate over the best baseline.

\subsection{Generalizability of UPTURN}
\label{subsec:transfer}

\begin{figure}[h]
    \centering
    \vspace{-1em}
    \begin{minipage}[t]{0.48\linewidth}
        \vspace{0pt}
        \centering
        {\captionsetup[subfigure]{skip=-1pt}\subcaptionbox{Position-policy transfer\label{fig:transfer-matrix}}[\linewidth]{%
            \includegraphics[width=\linewidth]{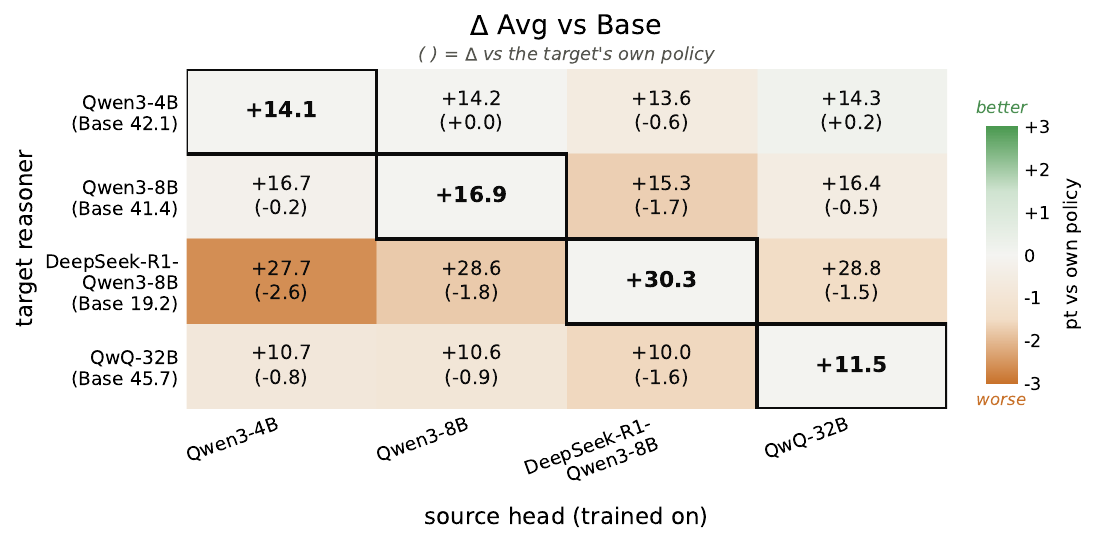}%
        }
        }
    \end{minipage}
    \hfill
    \begin{minipage}[t]{0.485\linewidth}
        \vspace{0pt}
        \centering
        {\captionsetup[subfigure]{skip=-1pt}\subcaptionbox{UPTURN gains over Base and GRPO\label{fig:grpo-incremental}}[\linewidth]{%
            \includegraphics[width=\linewidth]{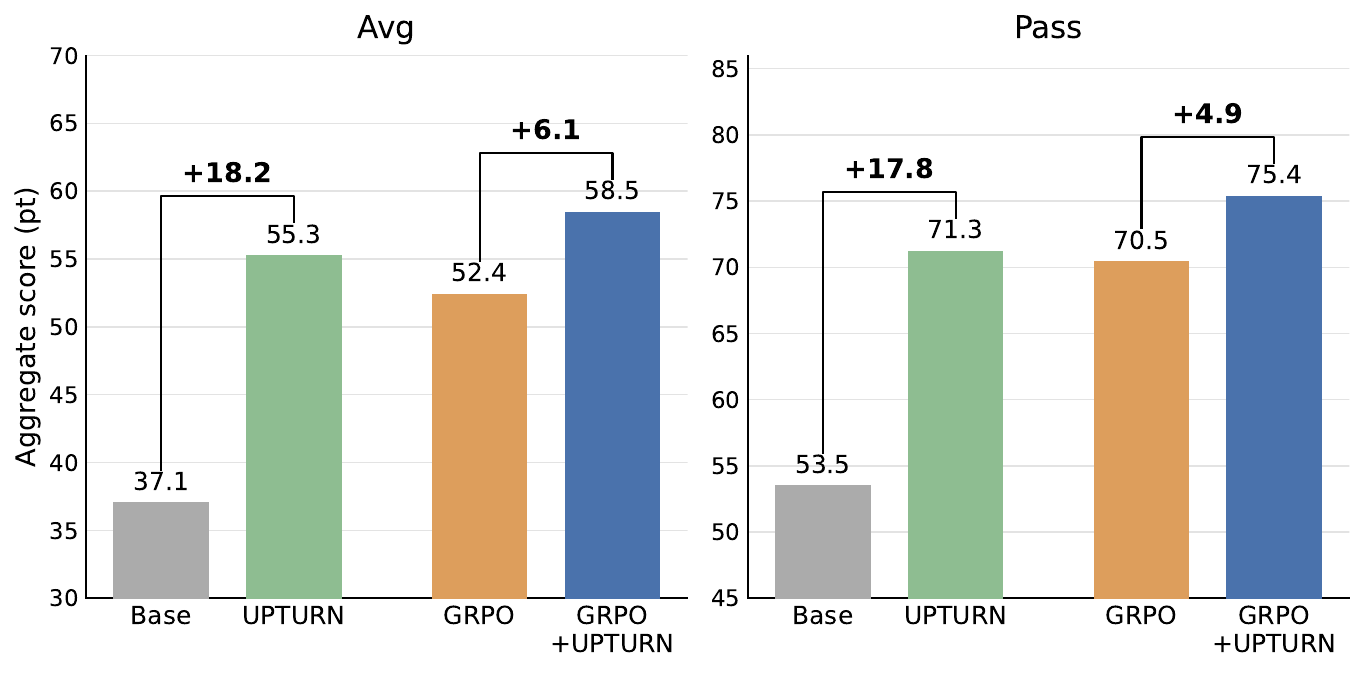}%
        }
        }
    \end{minipage}
    \vspace{-1em}
    \caption{\small{\textbf{Position-policy transfer.} Averages over six benchmarks. \textbf{(a)} Avg gain over Base when the policy trained on the source reasoner is applied to the target; parentheses (and colour) give the difference from the target's own policy on the diagonal. \textbf{(b)} UPTURN gains over Base and GRPO, averaged over four reasoners; the GRPO setting reuses the corresponding Base-trained position policy without retraining.}}
    \label{fig:transfer}
\end{figure}

\vspace{-.5em}

\paragraph{Transfer across reasoners.}
Because the position policy uses a frozen encoder rather than reasoner-specific hidden states, we transfer it while keeping the target reasoner, gate, and correction pipeline fixed. Transfer incurs little performance loss (Figure~\ref{fig:transfer-matrix}): averaged over all target--source pairs, transferred policies improve Avg over Base by $+17.2$ points, versus $+18.2$ for target-trained policies; no transfer falls more than $2.6$ points below its target-trained counterpart. Relative to the gain over Base, this small gap suggests that the policy remains effective when reused across reasoners.

\ifdefined\UPTURNstackedtables
    \begingroup
\else
\ifdefined\UPTURNcolm
\begin{wraptable}[10]{r}{0.5\linewidth}
    \vspace{-1.4em}
\else
\begin{wraptable}[9]{r}{0.5\linewidth}
    \vspace{-1.35em}
\fi
\fi
    \centering
    \setlength{\tabcolsep}{0.0098\linewidth}
    \renewcommand{\arraystretch}{1.0}
    \captionsetup{font=footnotesize}
    \caption{\textbf{Transfer to GPQA-Diamond}. $\Delta$ denotes UPTURN's gain over Base; \textbf{bold} marks the higher score. DS-R1 abbreviates DeepSeek-R1-Qwen3-8B.}
    \ifdefined\UPTURNstackedtables
        \vspace{0.1em}
    \else
        \vspace{-0.9em}
    \fi
    \fontsize{8pt}{8pt}\selectfont
    \begin{tabular}{l C{0.086\linewidth}C{0.086\linewidth} C{0.132\linewidth}C{0.132\linewidth} C{0.097\linewidth}C{0.097\linewidth}}
        \toprule
        \multirow{2}{*}{\textbf{Reasoner}}
            & \multicolumn{2}{c}{\textbf{Base@32}}
            & \multicolumn{2}{c}{\textbf{UPTURN@16}}
            & \multicolumn{2}{c}{$\boldsymbol{\Delta}$} \\
        \cmidrule(lr){2-3} \cmidrule(lr){4-5} \cmidrule(lr){6-7}
            & {\scriptsize Avg} & {\scriptsize Pass} & {\scriptsize Avg} & {\scriptsize Pass} & {\scriptsize Avg} & {\scriptsize Pass} \\
        \midrule
        Qwen3-4B    & 32.5 & 56.6 & \textbf{50.4} & \textbf{79.8} & \cellcolor{gray!10}\textcolor{deltacol}{+17.9} & \cellcolor{gray!10}\textcolor{deltacol}{+23.2} \\
        Qwen3-8B    & 35.0 & 61.1 & \textbf{54.3} & \textbf{83.8} & \cellcolor{gray!10}\textcolor{deltacol}{+19.3} & \cellcolor{gray!10}\textcolor{deltacol}{+22.7} \\
        DS-R1 & 20.7 & 49.0 & \textbf{35.9} & \textbf{78.8} & \cellcolor{gray!10}\textcolor{deltacol}{+15.2} & \cellcolor{gray!10}\textcolor{deltacol}{+29.8} \\
        QwQ-32B     & 41.5 & 68.7 & \textbf{54.2} & \textbf{80.8} & \cellcolor{gray!10}\textcolor{deltacol}{+12.7} & \cellcolor{gray!10}\textcolor{deltacol}{+12.1} \\
        \bottomrule
    \end{tabular}
    \label{tab:domain_transfer}
\ifdefined\UPTURNstackedtables
    \par\endgroup
\else
    \vspace{-1.2em}
    \end{wraptable}
\fi

\paragraph{Transfer to another domain.}
UPTURN has so far been trained and tested only in mathematics. To assess domain generalisation, we evaluate on GPQA-Diamond~\citep{rein2024gpqa}, a benchmark of 198 graduate-level multiple-choice questions in physics, chemistry, and biology, under the protocol of Table~\ref{tab:main_results} and without retraining, retuning, or changing the gate threshold on science data. Even so, UPTURN outperforms Base for every reasoner, as shown in Table~\ref{tab:domain_transfer}.

\ifdefined\UPTURNstackedtables
    \begingroup
\else
    \ifdefined\UPTURNcolm
    \begin{wraptable}[11]{r}{0.5\linewidth}
    \vspace{-0.4em}
    \else
    \begin{wraptable}[8]{r}{0.5\linewidth}
    \vspace{-1.3em}
    \fi
\fi
    \centering
    \setlength{\tabcolsep}{3pt}
    \renewcommand{\arraystretch}{1.0}
    \captionsetup{font=footnotesize}
    \caption{\textbf{UPTURN after self-correction.} Avg\,/\,Pass averaged over six benchmarks and three reasoners (excluding QwQ-32B), at $2L_{\max}$. $\Delta$ gives the added gain; \textbf{bold} marks the better score in each pair.}
    \label{tab:complementary_summary}
    \vspace{-0.8em}
    \fontsize{8pt}{8pt}\selectfont
    \begin{tabular*}{\linewidth}{@{\extracolsep{\fill}}l c c@{}}
        \toprule
        \textbf{Avg / Pass} & \textbf{Reflexion@16} & \textbf{SCoRe@16} \\
        \midrule
        Baseline & 55.3 / 77.5 & 52.3 / 79.9 \\
        + UPTURN & \textbf{64.1} / \textbf{80.5} & \textbf{64.1} / \textbf{81.8} \\
        $\Delta$ & \cellcolor{gray!10}\textcolor{deltacol}{+8.8 / +3.0} & \cellcolor{gray!10}\textcolor{deltacol}{+11.8 / +1.8} \\
        \bottomrule
    \end{tabular*}
\ifdefined\UPTURNstackedtables
    \par\endgroup
\else
    \vspace{-0.8em}
    \end{wraptable}
\fi

\paragraph{Complementarity with self-correction.}
We apply UPTURN once to gate-flagged solutions from each baseline's corrected pool, without retraining any component. Under a total response-token budget of $2L_{\max}$, UPTURN adds $+8.8$ Avg and $+3.0$ Pass points to Reflexion, and $+11.8$ Avg and $+1.8$ Pass points to SCoRe (Table~\ref{tab:complementary_summary}). These gains show that UPTURN also complements existing self-correction methods. Full results are in Appendix~\ref{app:complementary}.

\paragraph{Complementarity with GRPO.}
Figure~\ref{fig:grpo-incremental} tests whether UPTURN remains useful after reasoner post-training. Applying the corresponding Base-trained position policy to each GRPO-post-trained reasoner yields further gains of $+6.1$ Avg and $+4.9$ Pass points, averaged over four reasoners, without retraining the policy. These additional gains show that UPTURN complements GRPO.

Together, these results suggest that the learned position policy can support correction across reasoners, training stages, and domains.

\subsection{Ablation}
\label{subsec:ablation}

\begin{figure}[h]
    \vspace{-0.5em}
    \centering
    \captionsetup[subfigure]{skip=0pt}
    \begin{minipage}[t]{0.4525\linewidth}
        \vspace{0pt}
        \centering
        \subcaptionbox{Cut-position ablation\label{fig:ablation-position}}[\linewidth]{%
            \includegraphics[width=\linewidth]{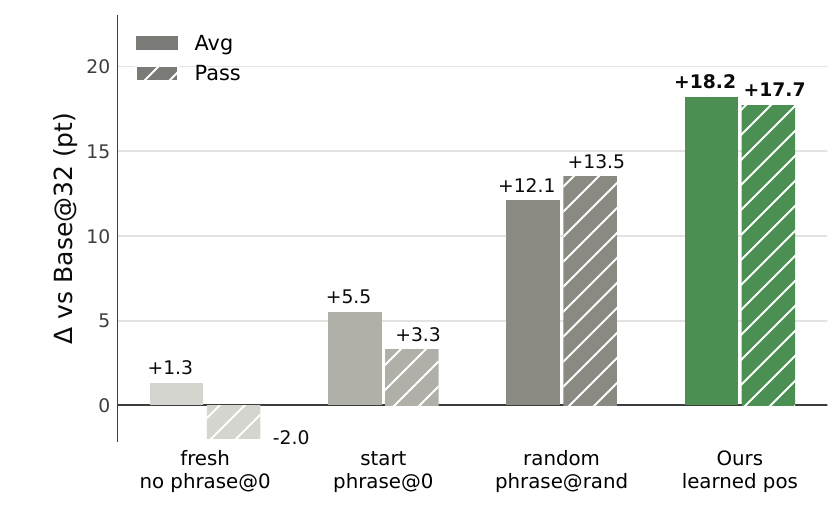}%
        }
    \end{minipage}
    \hfill
    \begin{minipage}[t]{0.5275\linewidth}
        \vspace{0pt}
        \centering
        \subcaptionbox{Inline vs. multi-turn injection\label{fig:ablation-injection}}[\linewidth]{%
            \includegraphics[width=\linewidth]{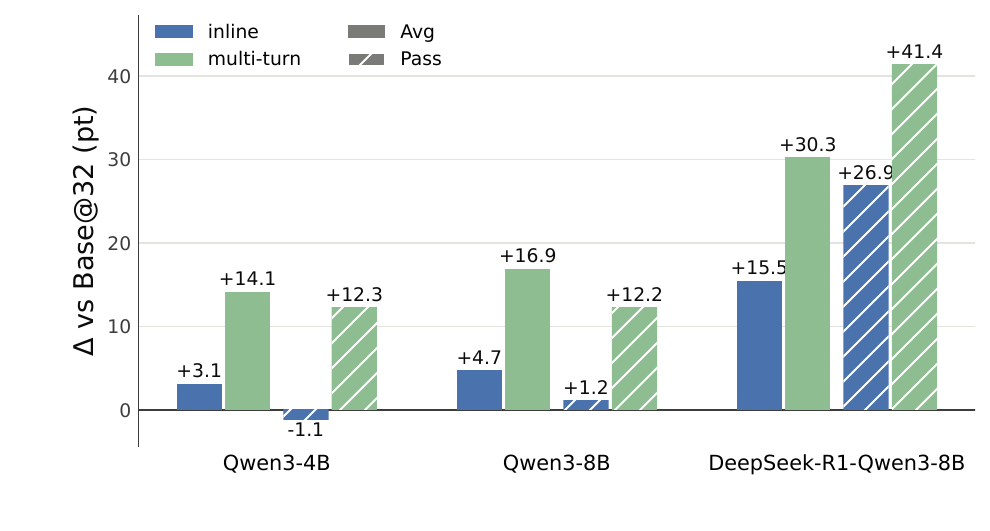}%
        }
    \end{minipage}
    \vspace{-0.5em}
    \caption{\small{\textbf{Position and injection ablations.} Avg and Pass gains over Base@32, averaged over six benchmarks. \textbf{(a)} Fresh regeneration and injection at start, random, or learned positions, using the same gate and replacement rule; averages over four reasoners. \textbf{(b)} Inline and multi-turn injection with separately trained position policies.
    }}
    \label{fig:ablations}
\end{figure}

\paragraph{Position: where to cut.}
All variants in Figure~\ref{fig:ablation-position} use the same gate and replacement rule but differ in how regeneration is initiated. \emph{Fresh} discards the flagged rollout and resamples it from the original prompt with the full budget $L_{\max}$; \emph{start} injects the instruction at the very beginning of the rollout, so no prefix is retained; \emph{random} injects it at a uniformly sampled paragraph boundary. Averaged over four reasoners and six benchmarks, the learned position performs best of the four variants.

\paragraph{Injection: learning the position for inline injection.}
One might suspect that inline injection fails in \S\ref{sec:motivation} because its position is not optimized. To test this possibility, we train a separate position policy for inline injection under exactly the training setup of \S\ref{subsec:training}, changing only the injection mode. As shown in Figure~\ref{fig:ablation-injection}, inline injection still yields smaller Avg and Pass gains than multi-turn injection for every reasoner. Thus, optimizing the injection position does not eliminate the gap, supporting the advantage of turn-boundary injection observed in \S\ref{sec:motivation}. Appendix~\ref{app:phrase} further examines sensitivity to instruction wording, including an empty user turn with no injected text.
We also examine whether additional correction rounds yield further gains (Appendix~\ref{app:iter-rounds}).

\ifdefined\UPTURNstackedtables
    \begingroup
\else
    \begin{wraptable}[10]{r}{0.48\linewidth}
    \vspace{-1.3em}
\fi
    \centering
    \setlength{\tabcolsep}{3pt}
    \renewcommand{\arraystretch}{1.0}
    \captionsetup{font=footnotesize}
    \caption{\textbf{Gate ablations}, averaged over four reasoners and six benchmarks. \textbf{GS}: accuracy of the highest-gate candidate. {+}GT: flagging with a ground-truth gate.}
    \label{tab:gate_select_summary}
    \vspace{-0.8em}
    \fontsize{8pt}{8pt}\selectfont
\begin{tabular*}{\linewidth}{@{\extracolsep{\fill}}l ccc@{}}
    \toprule
    \textbf{Method} & \textbf{Avg} & \textbf{GS} & \textbf{Pass} \\
    \midrule
    Base@32 & 37.1 & 52.0 & 53.5 \\
    \textbf{UPTURN@16} & \textbf{55.3} & \textbf{63.0} & \textbf{71.3} \\
    $\Delta$ & \cellcolor{gray!10}\textcolor{deltacol}{+18.2} & \cellcolor{gray!10}\textcolor{deltacol}{+11.0} & \cellcolor{gray!10}\textcolor{deltacol}{+17.7} \\
    \midrule
    UPTURN@16 & 55.3 & 63.0 & \textbf{71.3} \\
    UPTURN@16{+}GT & \textbf{55.9} & \textbf{63.3} & \textbf{71.3} \\
    $\Delta$ & \cellcolor{gray!10}\textcolor{deltacol}{+0.6} & \cellcolor{gray!10}\textcolor{deltacol}{+0.2} & \cellcolor{gray!10}\textcolor{deltacol}{+0.1} \\
    \bottomrule
\end{tabular*}

\ifdefined\UPTURNstackedtables
    \par\endgroup
\else
    \vspace{-0.8em}
    \end{wraptable}
\fi

\paragraph{Gate: selecting and flagging.}
We test the gate in two ways (Table~\ref{tab:gate_select_summary}). First, as a selector: we return the highest-gate candidate per question and report its accuracy (\textbf{GS}). Base@32 reaches $52.0$ GS, close to its $53.5$ Pass, so the gate reliably picks a correct rollout when one exists, and UPTURN@16 keeps a $+11.0$ point advantage under gate selection. Second, we flag with a ground-truth gate instead of the learned one. All metrics change by less than one point, so the learned gate is already close to optimal for flagging (details in Appendix~\ref{app:gate}).

\section{Conclusion}

We frame reasoning correction as redirecting an existing trajectory rather than retraining the reasoner. UPTURN builds on the observation that inline text is absorbed into the assistant's reasoning, whereas the same instruction delivered across a turn boundary can redirect it. A lightweight gate flags likely failures, a 130K-parameter head on a frozen encoder selects a cut that preserves a useful prefix, and the frozen reasoner regenerates from a fixed instruction injected as a new user turn.
Across four reasoners and six math benchmarks, UPTURN consistently improves average accuracy and outperforms self-correction and reflection-token injection baselines under matched correction budgets, as well as heuristic cuts and inline injection. It transfers across reasoners and to GPQA-Diamond without retraining, and provides additional gains on top of GRPO and existing self-correction methods. When the required capability is already present, a small learned intervention can thus recover a better reasoning path without retraining the reasoner; extension to code and tool use remains open.

\subsubsection*{Acknowledgments}
This work was supported by the Institute of Information \& Communications Technology Planning \& Evaluation (IITP) grant funded by the Korea government (MSIT) (No.RS-2022-II220124, Development of Artificial Intelligence Technology for Self-Improving Competency-Aware Learning Capabilities) and by the National Research Foundation of Korea (NRF) grant funded by the Korea government (MSIT) (No. RS-2025-00557589, Generative Model Based Efficient Reinforcement Learning Algorithms for Multi-modal Expansion in Generalized Environments), and by the “Advanced GPU Utilization Support Program” funded by the Government of the Republic of Korea (Ministry of Science and ICT).

\subsection*{AI use statement}
In this work, we used generative AI tools to assist with editing the manuscript (formatting tables, revising text for clarity, and updating reported numbers from experiment logs) and as a coding assistant for analysis and plotting scripts. All AI-assisted text, numbers, and code were reviewed and verified by the authors: reported results were checked against the original experiment outputs, and analysis code was validated by reproducing previously verified results. Generative AI was not used to generate research ideas, related-work surveys, or experimental results. We take responsibility for the final content of this work, including text, claims, and artifacts produced with the aid of generative AI.

\bibliographystyle{iclr2027_conference}
\bibliography{iclr2027_conference}

\clearpage
\appendix

\section{Illustrative Correction Example}
\label{app:motivation-example}

\begin{figure}[h]
    \centering
    \includegraphics[width=\linewidth]{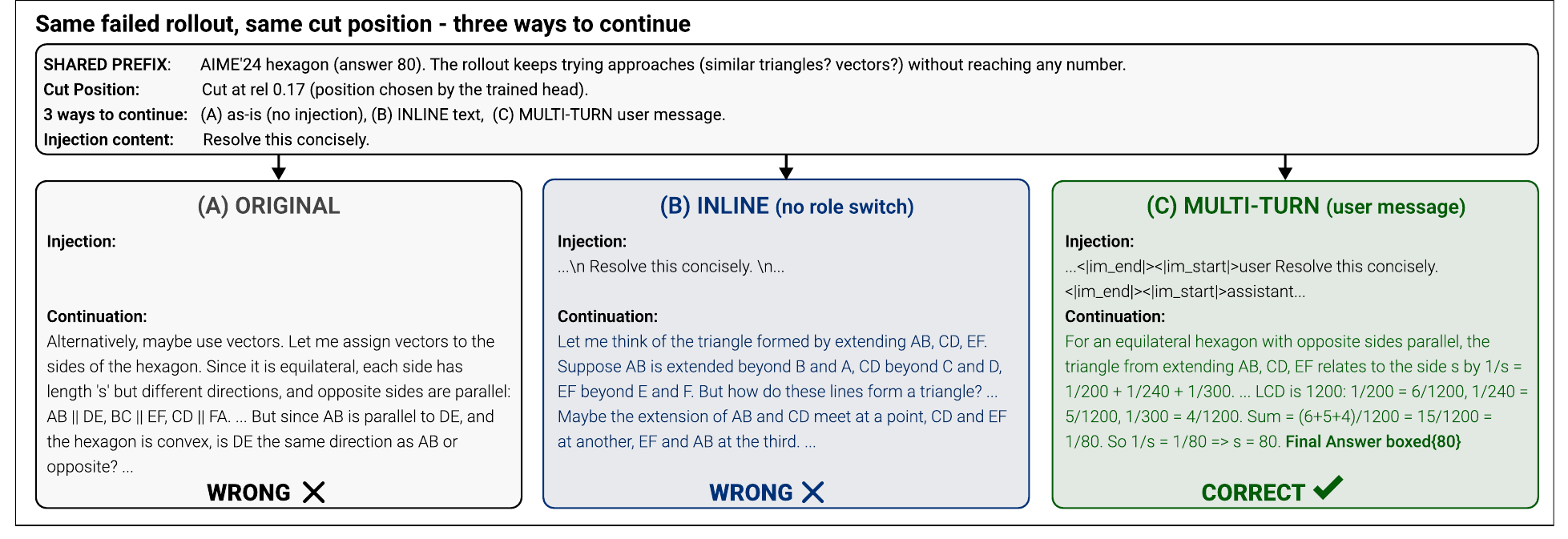}
    \caption{\textbf{One failed Qwen3-4B (Thinking on) rollout continued three ways from the same cut.} \textbf{(A)} Unmodified continuation remains trapped in re-verification. \textbf{(B)} Inline injection is absorbed as the model's own reasoning and resumes the loop. \textbf{(C)} The same instruction in a new user turn redirects reasoning to the correct answer. Full rollouts are provided in Appendix~\ref{app:full-rollouts}.}
    \label{fig:motivation}
\end{figure}

\section{Limitations}
\label{app:limitations}

UPTURN is a first step toward correcting reasoning by redirection rather than retraining, and several aspects remain open. Our study focuses on reasoning tasks whose outcomes can be checked automatically, and on a single, deliberately minimal form of intervention. Richer interventions, and settings where success is harder to verify, may call for different choices of what to inject and where. We leave these extensions to future work.

\section{Implementation Details}
\label{app:impl}

\paragraph{Position policy architecture.}
The position head $f_\phi$ operates on the 1024-dimensional last-layer hidden states of the frozen Qwen3-0.6B encoder. It applies LayerNorm to the input, followed by a linear layer $1024\!\to\!128$, a GELU activation, and a linear layer $128\!\to\!1$ that produces the logit of each candidate position, totaling 133{,}377 parameters (about 130K), the only trainable parameters of the position policy. The encoder receives exactly the token sequence the reasoner produced, the chat-templated prompt tokens concatenated with the rollout tokens, with no re-tokenization, as all our reasoners share the encoder's 151{,}643-token Qwen vocabulary. The delimiter \texttt{\textbackslash n\textbackslash n} is a single token in this tokenizer, and a failed rollout contains 11.3 candidate positions on average (median 5).

\paragraph{Training details.}
We train on DAPO-17K, a 17K-prompt subset of the DAPO-Math training pool. With dynamic sampling (at most 8 generation batches per step), each of the 30 training steps collects $B{=}64$ prompts containing at least one failed rollout, so training consumes 1{,}920 kept prompts in total. The head is updated once per step, sweeping all failed rollouts of the batch in mini-batches of 32 for one epoch, so the behavior policy $\pi_{\phi_{\text{old}}}$ is simply the policy that generated the step's data. Rollouts and regenerations use identical sampling settings. When a cut position falls inside an unclosed \texttt{<think>} block, we append \texttt{</think>} before injection so that the assistant turn remains well-formed, at both training and inference. Table~\ref{tab:hyperparams} lists all hyperparameters.

\begin{table}[h]
\centering
\caption{Hyperparameters}
\label{tab:hyperparams}
\begin{tabular}{ll}
\toprule
Hyperparameter & Value \\
\midrule
Batch size $B$ (prompts with $\ge$1 failure) & 64 \\
Rollouts per prompt $N$ & 8 \\
Cut positions per rollout $G$ & 8 \\
Continuations per position $M$ & 8 \\
Generation budget $L_{\max}$ & 8192 \\
Max generation batches (dynamic sampling) & 8 \\
Training steps & 30 \\
Mini-batch size / epochs per step & 32 / 1 \\
Optimizer & AdamW \\
Learning rate & $1\times10^{-3}$ (constant, no warmup) \\
Weight decay & 0 \\
Clipping range $\varepsilon$ & 0.2 \\
Entropy coefficient $\beta_{\mathrm{ent}}$ & 0.001 \\
Sampling (rollout and regeneration) & temp.\ 0.6, top-$p$ 0.95, top-$k$ 20 \\
\bottomrule
\end{tabular}
\end{table}

\paragraph{Inference settings.}
The gate threshold is $\gamma{=}0.4$; Appendix~\ref{app:sensitivity} shows that results are robust to it. Each flagged rollout is corrected once and replaced by its regeneration, with no iterative rounds. The cut position is sampled from $\pi_\phi(\cdot\mid q,o)$. 

\paragraph{Data separation.}
All evaluation benchmarks (AIME24/25/26, AMC23, HMMT, MATH500) are held-out test sets and do not overlap with the DAPO training pool. The gate's 1{,}197-prompt pool (Appendix~\ref{app:gate}) is an independently sampled, topic-balanced subset of the same DAPO training pool.

\paragraph{Compute and software.}
All experiments run on NVIDIA B200 GPUs. The stack is a fork of verl, with vLLM 0.15.1 for generation, transformers 4.57.1, and PyTorch 2.9.1.

\paragraph{Instruction.}
The fixed instruction is \colorbox{deltacol!15}{$c=$ \texttt{``Resolve this concisely.''}}, used for all rollouts at both training and inference.

\paragraph{Multi-turn injection across model families.}
The turn boundary is realized with each reasoner's own chat-template tokens; Table~\ref{tab:mt_format} lists the sequence injected at the cut for each family.

\begin{table}[h]
    \centering
    \small
    \renewcommand{\arraystretch}{1.1}
    \caption{\textbf{Turn-boundary injection sequence per model family.} \colorbox{deltacol!15}{$c$} denotes the fixed instruction.}
    \vspace{-.5em}
    \begin{tabular}{ll}
        \toprule
        \textbf{Family (reasoners)} & \textbf{Injected sequence at the cut} \\
        \midrule
        \begin{tabular}[c]{@{}l@{}}Qwen ChatML\\(Qwen3-4B/8B, QwQ-32B)\end{tabular} &
        \begin{tabular}[c]{@{}l@{}}\texttt{<|im\_end|>\textbackslash n<|im\_start|>user\textbackslash n} \colorbox{deltacol!15}{$c$}\\\texttt{<|im\_end|>\textbackslash n<|im\_start|>assistant\textbackslash n}\end{tabular} \\
        \addlinespace[3pt]
        \begin{tabular}[c]{@{}l@{}}DeepSeek\\(DeepSeek-R1-0528-Qwen3-8B)\end{tabular} &
        \begin{tabular}[c]{@{}l@{}}\texttt{<\textbar end\_of\_sentence\textbar><\textbar User\textbar>} \colorbox{deltacol!15}{$c$}\\\texttt{<\textbar Assistant\textbar><think>\textbackslash n}\end{tabular} \\
        \bottomrule
    \end{tabular}
    \label{tab:mt_format}
\end{table}

\clearpage
\section{Gate Details}
\label{app:gate}

\paragraph{Training.}
The gate is trained once and reused unchanged across all experiments. Its training data comes from a single rollout pass, independent of the correction policy: we roll out 1{,}197 DAPO prompts, an independently sampled subset of the training pool in Appendix~\ref{app:impl}, with Qwen3-4B-Instruct-2507 (16 samples per prompt), label each rollout with a valid final answer by correctness, and cache the hidden state $h_{\mathrm{box}}$ of the frozen encoder at the final \texttt{\textbackslash boxed\{\}} token. Class-balancing yields 2{,}874 examples, on which the gate MLP, a 66K-parameter network separate from the 130K position head, is fit in seconds with the binary cross-entropy loss
\begin{equation}
  \mathcal{L}_{\mathrm{BCE}}(\psi) = -\frac{1}{n}\sum_{j=1}^{n}\Big[y_j\log \hat{p}_j + (1-y_j)\log\big(1-\hat{p}_j\big)\Big],
  \qquad \hat{p}_j=\sigma\!\big(\mathrm{MLP}_\psi(h_{\mathrm{box},j})\big),
  \label{eq:gate-loss}
\end{equation}
where $y_j\in\{0,1\}$ is the correctness label. The temperature $\kappa$ in Eq.~\eqref{eq:gate} is calibrated post hoc on a held-out prompt split by minimizing BCE with the MLP weights frozen, giving $\kappa=1.38$. The same gate is reused for every reasoner in our experiments, including thinking-mode models of different scales, since its features come from the frozen encoder reading the question and rollout text and are not tied to the generating model. The only non-trivial cost is the single rollout pass, so the gate is effectively free to fit, and in practice never needs refitting.

\paragraph{Ranking quality.}
Before any correction is generated, Table~\ref{tab:gate_auc} measures how well the gate score ranks correct rollouts above incorrect ones, scored over the rollouts that produced an answer (a rollout that produces none is assigned gate $0$ by convention and is never a ranking decision the gate has to make). Across the four reasoners the gate reaches a mean AUC of $.962$, with benchmark-wise AUC remaining at least $.920$.

\begin{table}[!ht]
    \centering
    \renewcommand{\arraystretch}{1.0}
    \caption{\textbf{Gate AUC by benchmark.} Mean over four reasoners.}
    \vspace{-.5em}
    \begin{tabular}{lccccccc}
        \toprule
        \textbf{Benchmark} & AIME24 & AIME25 & AIME26 & AMC23 & HMMT & MATH500 & \textbf{Avg} \\
        \midrule
        \textbf{AUC} & .982 & .983 & .980 & .984 & .920 & .922 & \textbf{.962} \\
        \bottomrule
    \end{tabular}
    \label{tab:gate_auc}
\end{table}

\paragraph{Learned versus ground-truth gating.}
\label{app:gate-oracle}
The gate decides only \emph{which} rollouts are corrected; it does not choose the cut, write the instruction, or regenerate. To see how much the reported numbers depend on that decision, we replace it with an oracle. In the ground-truth condition ({+}GT), a rollout is flagged exactly when its answer is wrong, instead of by the learned gate. Everything else is held fixed, so the oracle bounds what a perfect gate could buy.

Table~\ref{tab:metric_summary} reports the comparison. Averaged over all four reasoners, the ground-truth gate changes Avg only from $55.3$ to $55.9$ and leaves Pass at $71.3$. This small gap shows that the lightweight learned gate already identifies which rollouts require correction reliably, leaving little room for an oracle gate to improve the final result.

\begin{table}[!ht]
    \centering
    \setlength{\tabcolsep}{3pt}
    \renewcommand{\arraystretch}{1.0}
    \caption{\textbf{Learned (UPTURN@16) versus ground-truth ({+}GT) gating}, averaged over six benchmarks. \textbf{Bold} marks the higher score; ties are both bold.}
    \vspace{-.5em}
    \begin{tabular}{l ccc c@{} ccc}
        \toprule
        \multirow{2.4}{*}{\textbf{Model}}
        & \multicolumn{3}{c}{\textbf{Avg}} &
        & \multicolumn{3}{c}{\textbf{Pass}} \\
        \cmidrule(lr){2-4} \cmidrule(lr){6-8}
        & {\scriptsize Base@32} & {\scriptsize UPTURN@16} & {\scriptsize UPTURN@16{+}GT} &
        & {\scriptsize Base@32} & {\scriptsize UPTURN@16} & {\scriptsize UPTURN@16{+}GT} \\
        \midrule
        Qwen3-4B    & 42.1 & 56.2 & \textbf{56.6}          && 61.1 & 73.4          & \textbf{73.5} \\
        Qwen3-8B    & 41.4 & 58.3 & \textbf{58.7}          && 60.5 & 72.7          & \textbf{72.8} \\
        DeepSeek-R1-Qwen3-8B & 19.2 & 49.5 & \textbf{50.9}          && 28.8 & \textbf{70.3}          & \textbf{70.3} \\
        QwQ-32B     & 45.7 & 57.2 & \textbf{57.4}          && 63.8 & \textbf{68.7}          & \textbf{68.7} \\
        \midrule
        Model Avg   & 37.1 & 55.3          & \textbf{55.9} && 53.5 & \textbf{71.3}          & \textbf{71.3} \\
        \bottomrule
    \end{tabular}
    \label{tab:metric_summary}
\end{table}

\clearpage
\paragraph{Sensitivity to $\gamma$.}
\label{app:sensitivity}
Table~\ref{tab:sensitivity} sweeps $\gamma\in[0,1]$ over four reasoners and six benchmarks. At $\gamma{=}0$ no rollout is corrected and the pool is the uncorrected 16 rollouts; from $\gamma{=}0.1$ on, Avg varies by at most $0.4$ points and Pass by at most $0.1$, so the reported gains do not depend on tuning $\gamma$. Because gate scores concentrate near the extremes (mean AUC $.962$, Table~\ref{tab:gate_auc}), raising $\gamma$ mainly sends already-confident rollouts to a redundant correction: the fraction corrected grows from $64\%$ at $\gamma{=}0.1$ to $100\%$ at $\gamma{=}1.0$ while accuracy stays essentially flat. We therefore set $\gamma{=}0.4$ as a cost--coverage compromise rather than by tuning for accuracy.

\begin{table}[h]
    \centering
    \setlength{\tabcolsep}{7pt}
    \renewcommand{\arraystretch}{1.05}
    \caption{\textbf{Sensitivity to the gate threshold $\gamma$.} Avg and Pass over four reasoners and six benchmarks; Corrected is the fraction of rollouts that receive a correction. \textbf{Bold} marks the default. At $\gamma{=}0$ nothing is corrected.}
    \vspace{-.5em}
    \begin{tabular}{lccccccc}
        \toprule
        $\gamma$ & 0 & 0.1 & 0.2 & \textbf{0.4} & 0.6 & 0.8 & 1.0 \\
        \midrule
        Avg            & 37.3 & 55.6 & 55.5 & \textbf{55.3} & 55.2 & 55.2 & 55.2 \\
        Pass           & 51.8 & 71.2 & 71.3 & \textbf{71.3} & 71.3 & 71.3 & 71.3 \\
        Corrected (\%) & 0 & 64 & 69 & \textbf{79} & 91 & 99 & 100 \\
        \bottomrule
    \end{tabular}
    \label{tab:sensitivity}
\end{table}

\paragraph{Gate as a selector.}
\label{app:gate-select}
Table~\ref{tab:gate_select} gives the per-benchmark results behind Table~\ref{tab:gate_select_summary}. Base@32 selects among its 32 rollouts and UPTURN@16 among the 16 candidates after correction, so both stay within the budget of Table~\ref{tab:main_results}; rollouts without an answer are never selected. UPTURN's GS gain over Base@32 is smaller than its Pass gain because corrected candidates are not always the highest-scored one, but UPTURN is higher in 21 of 24 cells, ties on AMC23 for both Qwen3 reasoners, and is lower only on HMMT for QwQ-32B.

\begin{table}[!ht]
    \centering
    \setlength{\tabcolsep}{3pt}
    \renewcommand{\arraystretch}{0.85}
    \footnotesize
    \caption{\textbf{Gate as a selector}, per benchmark. \textbf{GS} is the accuracy of the highest-gate candidate; \textbf{Avg} and \textbf{Pass} are as in Table~\ref{tab:main_results}. $\Delta$ denotes UPTURN's gain over Base; \textbf{bold} marks the higher score.}
    \label{tab:gate_select}
    \vspace{-0.5em}
    \begin{tabular}{p{2.4cm} C{0.92cm}C{0.92cm}C{0.92cm} C{3pt} C{0.92cm}C{0.92cm}C{0.92cm} C{3pt} C{0.92cm}C{0.92cm}C{0.92cm}}
        \toprule
        \multirow{2}{*}{\textbf{Method}} & \multicolumn{3}{c}{\textbf{AIME24}} & & \multicolumn{3}{c}{\textbf{AIME25}} & & \multicolumn{3}{c}{\textbf{AIME26}} \\
        \cmidrule(lr){2-4} \cmidrule(lr){6-8} \cmidrule(lr){10-12}
             & {\scriptsize Avg} & {\scriptsize GS} & {\scriptsize Pass} & & {\scriptsize Avg} & {\scriptsize GS} & {\scriptsize Pass} & & {\scriptsize Avg} & {\scriptsize GS} & {\scriptsize Pass} \\
        \midrule
        \multicolumn{12}{@{}l}{\textit{Qwen3-4B}} \\
        \quad Base@32 & 32.6 & 60.0 & 63.3 &  & 21.8 & 46.7 & 46.7 &  & 26.7 & 40.0 & 43.3 \\
        \quad \textbf{UPTURN@16} & \textbf{50.8} & \textbf{66.7} & \textbf{83.3} &  & \textbf{39.0} & \textbf{53.3} & \textbf{56.7} &  & \textbf{44.2} & \textbf{53.3} & \textbf{70.0} \\
        \quad $\Delta$ & \cellcolor{gray!10}\textcolor{deltacol}{+18.2} & \cellcolor{gray!10}\textcolor{deltacol}{+6.7} & \cellcolor{gray!10}\textcolor{deltacol}{+20.0} & \cellcolor{gray!10} & \cellcolor{gray!10}\textcolor{deltacol}{+17.2} & \cellcolor{gray!10}\textcolor{deltacol}{+6.7} & \cellcolor{gray!10}\textcolor{deltacol}{+10.0} & \cellcolor{gray!10} & \cellcolor{gray!10}\textcolor{deltacol}{+17.5} & \cellcolor{gray!10}\textcolor{deltacol}{+13.3} & \cellcolor{gray!10}\textcolor{deltacol}{+26.7} \\
        \addlinespace[1pt]
        \multicolumn{12}{@{}l}{\textit{Qwen3-8B}} \\
        \quad Base@32 & 35.8 & 56.7 & 56.7 &  & 20.9 & 43.3 & 43.3 &  & 24.7 & 46.7 & 50.0 \\
        \quad \textbf{UPTURN@16} & \textbf{53.8} & \textbf{60.0} & \textbf{70.0} &  & \textbf{39.2} & \textbf{46.7} & \textbf{56.7} &  & \textbf{47.7} & \textbf{63.3} & \textbf{66.7} \\
        \quad $\Delta$ & \cellcolor{gray!10}\textcolor{deltacol}{+17.9} & \cellcolor{gray!10}\textcolor{deltacol}{+3.3} & \cellcolor{gray!10}\textcolor{deltacol}{+13.3} & \cellcolor{gray!10} & \cellcolor{gray!10}\textcolor{deltacol}{+18.2} & \cellcolor{gray!10}\textcolor{deltacol}{+3.3} & \cellcolor{gray!10}\textcolor{deltacol}{+13.3} & \cellcolor{gray!10} & \cellcolor{gray!10}\textcolor{deltacol}{+23.0} & \cellcolor{gray!10}\textcolor{deltacol}{+16.7} & \cellcolor{gray!10}\textcolor{deltacol}{+16.7} \\
        \addlinespace[1pt]
        \multicolumn{12}{@{}l}{\textit{DeepSeek-R1-Qwen3-8B}} \\
        \quad Base@32 & 5.8 & 16.7 & 16.7 &  & 7.1 & 10.0 & 13.3 &  & 3.5 & 6.7 & 6.7 \\
        \quad \textbf{UPTURN@16} & \textbf{43.8} & \textbf{40.0} & \textbf{70.0} &  & \textbf{31.2} & \textbf{36.7} & \textbf{53.3} &  & \textbf{35.2} & \textbf{50.0} & \textbf{56.7} \\
        \quad $\Delta$ & \cellcolor{gray!10}\textcolor{deltacol}{+37.9} & \cellcolor{gray!10}\textcolor{deltacol}{+23.3} & \cellcolor{gray!10}\textcolor{deltacol}{+53.3} & \cellcolor{gray!10} & \cellcolor{gray!10}\textcolor{deltacol}{+24.2} & \cellcolor{gray!10}\textcolor{deltacol}{+26.7} & \cellcolor{gray!10}\textcolor{deltacol}{+40.0} & \cellcolor{gray!10} & \cellcolor{gray!10}\textcolor{deltacol}{+31.7} & \cellcolor{gray!10}\textcolor{deltacol}{+43.3} & \cellcolor{gray!10}\textcolor{deltacol}{+50.0} \\
        \addlinespace[1pt]
        \multicolumn{12}{@{}l}{\textit{QwQ-32B}} \\
        \quad Base@32 & 36.7 & 60.0 & 63.3 &  & 26.2 & 43.3 & 46.7 &  & 27.4 & 56.7 & 56.7 \\
        \quad \textbf{UPTURN@16} & \textbf{54.4} & \textbf{66.7} & \textbf{70.0} &  & \textbf{33.8} & \textbf{53.3} & \textbf{56.7} &  & \textbf{49.4} & \textbf{60.0} & \textbf{63.3} \\
        \quad $\Delta$ & \cellcolor{gray!10}\textcolor{deltacol}{+17.7} & \cellcolor{gray!10}\textcolor{deltacol}{+6.7} & \cellcolor{gray!10}\textcolor{deltacol}{+6.7} & \cellcolor{gray!10} & \cellcolor{gray!10}\textcolor{deltacol}{+7.5} & \cellcolor{gray!10}\textcolor{deltacol}{+10.0} & \cellcolor{gray!10}\textcolor{deltacol}{+10.0} & \cellcolor{gray!10} & \cellcolor{gray!10}\textcolor{deltacol}{+22.0} & \cellcolor{gray!10}\textcolor{deltacol}{+3.3} & \cellcolor{gray!10}\textcolor{deltacol}{+6.7} \\
        \midrule\midrule
        \multirow{2}{*}{\textbf{Method}} & \multicolumn{3}{c}{\textbf{AMC23}} & & \multicolumn{3}{c}{\textbf{HMMT}} & & \multicolumn{3}{c}{\textbf{MATH500}} \\
        \cmidrule(lr){2-4} \cmidrule(lr){6-8} \cmidrule(lr){10-12}
             & {\scriptsize Avg} & {\scriptsize GS} & {\scriptsize Pass} & & {\scriptsize Avg} & {\scriptsize GS} & {\scriptsize Pass} & & {\scriptsize Avg} & {\scriptsize GS} & {\scriptsize Pass} \\
        \midrule
        \multicolumn{12}{@{}l}{\textit{Qwen3-4B}} \\
        \quad Base@32 & 67.0 & 92.5 & 92.5 &  & 19.2 & 27.3 & 27.3 &  & 85.0 & 91.4 & 93.2 \\
        \quad \textbf{UPTURN@16} & \textbf{84.2} & 92.5 & \textbf{95.0} &  & \textbf{28.0} & \textbf{33.3} & \textbf{39.4} &  & \textbf{91.0} & \textbf{92.4} & \textbf{95.8} \\
        \quad $\Delta$ & \cellcolor{gray!10}\textcolor{deltacol}{+17.2} & \cellcolor{gray!10}{0.0} & \cellcolor{gray!10}\textcolor{deltacol}{+2.5} & \cellcolor{gray!10} & \cellcolor{gray!10}\textcolor{deltacol}{+8.8} & \cellcolor{gray!10}\textcolor{deltacol}{+6.1} & \cellcolor{gray!10}\textcolor{deltacol}{+12.1} & \cellcolor{gray!10} & \cellcolor{gray!10}\textcolor{deltacol}{+6.0} & \cellcolor{gray!10}\textcolor{deltacol}{+1.0} & \cellcolor{gray!10}\textcolor{deltacol}{+2.6} \\
        \addlinespace[1pt]
        \multicolumn{12}{@{}l}{\textit{Qwen3-8B}} \\
        \quad Base@32 & 64.8 & 92.5 & 92.5 &  & 19.0 & 27.3 & 27.3 &  & 83.1 & 91.4 & 93.0 \\
        \quad \textbf{UPTURN@16} & \textbf{85.3} & 92.5 & \textbf{95.0} &  & \textbf{32.2} & \textbf{36.4} & \textbf{51.5} &  & \textbf{92.0} & \textbf{92.0} & \textbf{96.4} \\
        \quad $\Delta$ & \cellcolor{gray!10}\textcolor{deltacol}{+20.5} & \cellcolor{gray!10}{0.0} & \cellcolor{gray!10}\textcolor{deltacol}{+2.5} & \cellcolor{gray!10} & \cellcolor{gray!10}\textcolor{deltacol}{+13.2} & \cellcolor{gray!10}\textcolor{deltacol}{+9.1} & \cellcolor{gray!10}\textcolor{deltacol}{+24.2} & \cellcolor{gray!10} & \cellcolor{gray!10}\textcolor{deltacol}{+8.8} & \cellcolor{gray!10}\textcolor{deltacol}{+0.6} & \cellcolor{gray!10}\textcolor{deltacol}{+3.4} \\
        \addlinespace[1pt]
        \multicolumn{12}{@{}l}{\textit{DeepSeek-R1-Qwen3-8B}} \\
        \quad Base@32 & 32.3 & 47.5 & 47.5 &  & 2.0 & 3.0 & 9.1 &  & 64.5 & 75.6 & 79.8 \\
        \quad \textbf{UPTURN@16} & \textbf{79.5} & \textbf{90.0} & \textbf{97.5} &  & \textbf{24.6} & \textbf{27.3} & \textbf{48.5} &  & \textbf{82.8} & \textbf{89.8} & \textbf{95.6} \\
        \quad $\Delta$ & \cellcolor{gray!10}\textcolor{deltacol}{+47.2} & \cellcolor{gray!10}\textcolor{deltacol}{+42.5} & \cellcolor{gray!10}\textcolor{deltacol}{+50.0} & \cellcolor{gray!10} & \cellcolor{gray!10}\textcolor{deltacol}{+22.6} & \cellcolor{gray!10}\textcolor{deltacol}{+24.2} & \cellcolor{gray!10}\textcolor{deltacol}{+39.4} & \cellcolor{gray!10} & \cellcolor{gray!10}\textcolor{deltacol}{+18.3} & \cellcolor{gray!10}\textcolor{deltacol}{+14.2} & \cellcolor{gray!10}\textcolor{deltacol}{+15.8} \\
        \addlinespace[1pt]
        \multicolumn{12}{@{}l}{\textit{QwQ-32B}} \\
        \quad Base@32 & 73.0 & 87.5 & 87.5 &  & 22.7 & \textbf{33.3} & 33.3 &  & 87.8 & 93.0 & 95.0 \\
        \quad \textbf{UPTURN@16} & \textbf{86.2} & \textbf{92.5} & \textbf{92.5} &  & \textbf{27.1} & 30.3 & 33.3 &  & \textbf{92.2} & \textbf{93.6} & \textbf{96.2} \\
        \quad $\Delta$ & \cellcolor{gray!10}\textcolor{deltacol}{+13.2} & \cellcolor{gray!10}\textcolor{deltacol}{+5.0} & \cellcolor{gray!10}\textcolor{deltacol}{+5.0} & \cellcolor{gray!10} & \cellcolor{gray!10}\textcolor{deltacol}{+4.4} & \cellcolor{gray!10}\textcolor{red}{-3.0} & \cellcolor{gray!10}{0.0} & \cellcolor{gray!10} & \cellcolor{gray!10}\textcolor{deltacol}{+4.3} & \cellcolor{gray!10}\textcolor{deltacol}{+0.6} & \cellcolor{gray!10}\textcolor{deltacol}{+1.2} \\
        \bottomrule
    \end{tabular}
\end{table}

\clearpage
\section{Wall-Clock Cost}
\label{app:throughput}

The fixed-budget protocol controls generation budgets, but does not guarantee equal wall-clock time. UPTURN requires two dependent generation stages, gate evaluation of all initial rollouts, and position selection and retained-prefix processing for flagged rollouts. These extra steps may offset the time saved by decoding fewer tokens. We therefore time both pipelines end to end on one idle NVIDIA B200 with Qwen3-4B on AIME24 at $L_{\max}{=}8192$. Each pipeline is executed twice; sampling is seeded, so generation and token counts are identical across the two repetitions. We report wall-clock time as the mean $\pm$ half the absolute difference between the two measurements. Base@32 generates all 32 rollouts per question in a single $n{=}32$ batch. Table~\ref{tab:throughput} reports generated calls and decoded tokens per question and total wall-clock time for the benchmark, excluding the same 32\,s engine start-up cost for every method.

\begin{table}[!ht]
\centering
\caption{\textbf{Wall-clock cost on AIME24} (Qwen3-4B, one B200).}
\label{tab:throughput}
\begin{tabular}{lrrrr}
\toprule
Method & Gens./q & Gen.\ tokens/q & Wall-clock (s) & vs.\ Base@32 \\
\midrule
Base@16 & 16.0 & 122,881 & $477\pm2$ & $0.51\times$ \\
\rowcolor{gray!10}
Base@32 & 32.0 & 245,548 & $931\pm0$ & $1.00\times$ \\
\textbf{UPTURN@16} & \textbf{28.6} & \textbf{156,619} & $\mathbf{790\pm2}$ & $\mathbf{0.85\times}$ \\
\bottomrule
\end{tabular}
\end{table}

UPTURN reduces decoded tokens by approximately 36\% and wall-clock time by 15\% relative to Base@32 as shown in Table~\ref{tab:throughput}. 
The wall-clock gain is smaller than the token gain because the overhead of additional gating, position-selection, and prefix-processing. However, these costs are smaller than the decoding time UPTURN saves by generating fewer tokens, so it remains faster overall.

\clearpage
\section{Does a Second Correction Round Help?}
\label{app:iter-rounds}

UPTURN corrects each flagged rollout only once. We examine whether further rounds improve performance on Qwen3-4B and Qwen3-8B across all six benchmarks, using up to three correction rounds under two rules. \emph{Gated} accepts a regeneration only if it raises the gate score and stops correcting a rollout once its answer repeats; \emph{naive} always replaces the current rollout and applies all three rounds. Here, \emph{gated} refers to the acceptance and stopping rules, not the threshold used to select rollouts for correction. Both variants start from the same cached 16-rollout pool of Table~\ref{tab:main_results}, so round~0 is identical. We set $\gamma{=}1.0$ to remove threshold-based selection: all rollouts are initially eligible for correction, while subsequent rounds follow each variant's stopping rule. Because this protocol differs from the main evaluation, we focus on changes across rounds rather than absolute comparisons with Table~\ref{tab:main_results}.

\begin{figure}[h]
    \centering
    \includegraphics[width=\linewidth]{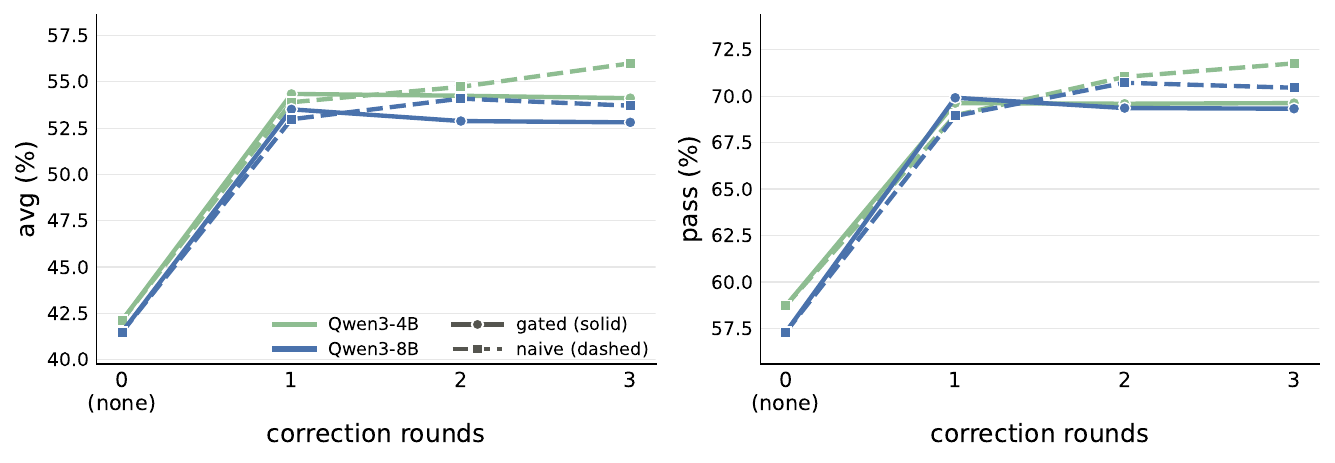}
    \caption{\textbf{Avg and Pass against the number of correction rounds}, averaged over six benchmarks. Colour marks the reasoner; solid lines are \emph{gated}, dashed lines \emph{naive}.}
    \label{fig:iter_rounds}
\end{figure}

Figure~\ref{fig:iter_rounds} shows that the first correction accounts for most of the improvement under both rules. Under \emph{gated} acceptance, additional rounds provide no further gains and slightly reduce Avg or Pass in some settings. At round~1, this rule achieves higher Avg and Pass than \emph{naive} replacement on both reasoners. With further naive rounds, performance continues to improve on Qwen3-4B, while Qwen3-8B peaks at round~2 and declines slightly at round~3, remaining above round~1 on both metrics. Overall, additional rounds require further generation but offer smaller, rule- and reasoner-dependent benefits, motivating our single-round default.

\clearpage
\section{Sensitivity to the Injection Instruction}
\label{app:phrase}

UPTURN injects a fixed instruction, ``\texttt{Resolve this concisely.}'', at the chosen cut. To separate the effect of the turn-boundary structure from that of the wording, we rerun the deployed pipeline on all six benchmarks varying only the injected text; the phase-1 pool, gate, position policy, budget, and seed are identical, so only the continuation after the cut differs. We compare four task-relevant instructions and three controls: the reflection token ``\texttt{Wait.}'', the meaningless ``\texttt{Turtle.}'', and \emph{None}, an empty user turn in which only the chat-template turn-boundary tokens are inserted at the cut with no text at all.

\begin{figure}[h]
    \centering
    \includegraphics[width=\linewidth]{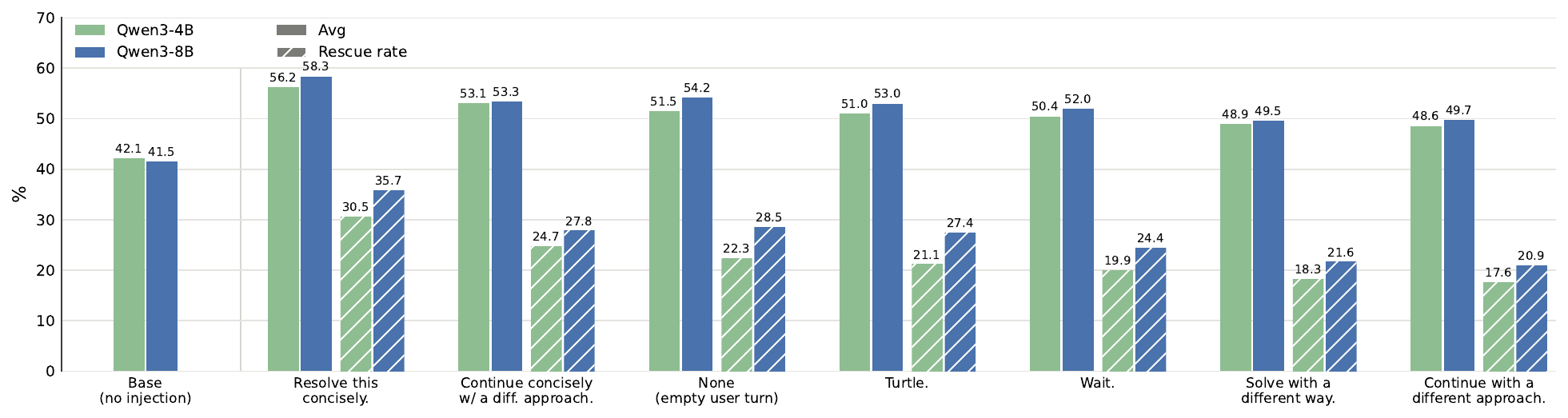}
    \caption{\textbf{Avg and rescue rate under varied injection instructions}, averaged over six benchmarks. Solid bars are Avg, hatched bars rescue rate; Base has no rescue rate. Instructions are ordered by mean Avg; \emph{None} inserts the turn boundary with no text.}
    \label{fig:phrase}
\end{figure}

Figure~\ref{fig:phrase} shows two effects. First, the injection structure matters most: every multi-turn variant improves Avg over Base on both reasoners, and even the empty user turn (\emph{None}) reaches $51.5$/$54.2$ Avg on Qwen3-4B/8B against $42.1$/$41.5$ for Base. This is consistent with Figure~\ref{fig:ablation-injection}, where multi-turn injection outperforms inline injection by a wide margin on every reasoner. Second, once the injection is multi-turn, the wording matters: Avg across the seven variants spans $48.6$ to $56.2$ on Qwen3-4B and $49.5$ to $58.3$ on Qwen3-8B, and the default instruction improves Avg over \emph{None} by $4.7$ and $4.1$ points, achieving the highest Avg and rescue rate on both reasoners. The turn boundary thus provides the structure that makes correction effective, and the instruction placed at it determines how much of that gain is realized.


\clearpage
\section{Paragraph Boundaries are Forking Points}
\label{app:entropy}

UPTURN restricts candidate cuts to \texttt{\textbackslash n\textbackslash n} boundaries. To check that this does not miss where a trace actually forks, we compute the next-token entropy at every position of the incorrect rollouts from the 16-rollout pool of Table~\ref{tab:main_results} on AIME24/25, using each reasoner itself. Table~\ref{tab:entropy_fork} shows the same pattern for all four reasoners. Entropy is consistently high right after a paragraph boundary and near zero at a typical position, so every boundary is a forking point. Boundaries are only $2$--$3\%$ of all positions (\emph{boundary frac.}), yet they cover $18$--$35\%$ of the most uncertain positions, the top-5\% by entropy (\emph{top-5\% coverage}).

\begin{table}[h]
    \centering
    \small
    \renewcommand{\arraystretch}{1.0}
    \caption{\textbf{Next-token entropy at \texttt{\textbackslash n\textbackslash n} boundaries vs.\ elsewhere} on incorrect AIME24/25 rollouts; coverage is the share of top-5\% entropy positions within $\pm2$ tokens of a boundary.}
    \vspace{-.5em}
    \begin{tabular}{l cc cc c c}
        \toprule
        & \multicolumn{2}{c}{\textbf{Boundary entropy}} & \multicolumn{2}{c}{\textbf{Other entropy}} & \textbf{Top-5\%} & \textbf{Boundary} \\
        \cmidrule(lr){2-3} \cmidrule(lr){4-5}
        \textbf{Reasoner} & mean & median & mean & median & \textbf{coverage} & \textbf{frac.} \\
        \midrule
        Qwen3-4B    & \textbf{0.62} & \textbf{0.61} & 0.29 & 0.008 & 18\% & 2.0\% \\
        Qwen3-8B    & \textbf{0.60} & \textbf{0.59} & 0.27 & 0.004 & 19\% & 2.0\% \\
        DeepSeek-R1-Qwen3-8B & \textbf{0.99} & \textbf{1.01} & 0.31 & 0.018 & 35\% & 3.2\% \\
        QwQ-32B     & \textbf{0.81} & \textbf{0.80} & 0.41 & 0.095 & 22\% & 2.6\% \\
        \bottomrule
    \end{tabular}
    \label{tab:entropy_fork}
\end{table}

A candidate set chosen by entropy would therefore contain the boundaries plus some extra positions inside paragraphs, while giving up two practical advantages: a \texttt{\textbackslash n\textbackslash n} cut always ends at a complete paragraph, so the turn boundary is grafted onto a well-formed prefix, whereas an entropy-chosen cut can fall in the middle of a sentence, and \texttt{\textbackslash n\textbackslash n} positions are found by string matching, whereas entropy-based candidates require an extra forward pass of the reasoner over every rollout at inference time. We therefore use \texttt{\textbackslash n\textbackslash n} boundaries as the candidate set of the position policy.

\clearpage
\section{Scaling the Generation Budget}
\label{app:lmax-scale}

Figure~\ref{fig:lmax-scale} extends the comparison of Table~\ref{tab:main_results} to per-rollout budgets from $L_{\max}$ to $3L_{\max}$ ($8$K to $24$K tokens) on Qwen3-4B, Qwen3-8B, and DeepSeek-R1-Qwen3-8B across six benchmarks. Base@32 draws 32 rollouts at each budget, while UPTURN@16 starts from 16 rollouts and corrects each gate-flagged one once. Both methods allow up to 32 generations per question and use the same per-rollout length limit; for UPTURN, the retained prefix and regenerated suffix together remain within that limit. These constraints match the allowed budgets, not actual token usage. We apply the position policy trained at $L_{\max}$ unchanged at every budget, without retraining or adjusting the gate threshold, to evaluate its transfer to longer generation windows. Policies retrained for larger budgets are not evaluated here.

\begin{figure}[h]
    \centering
    \includegraphics[width=\linewidth]{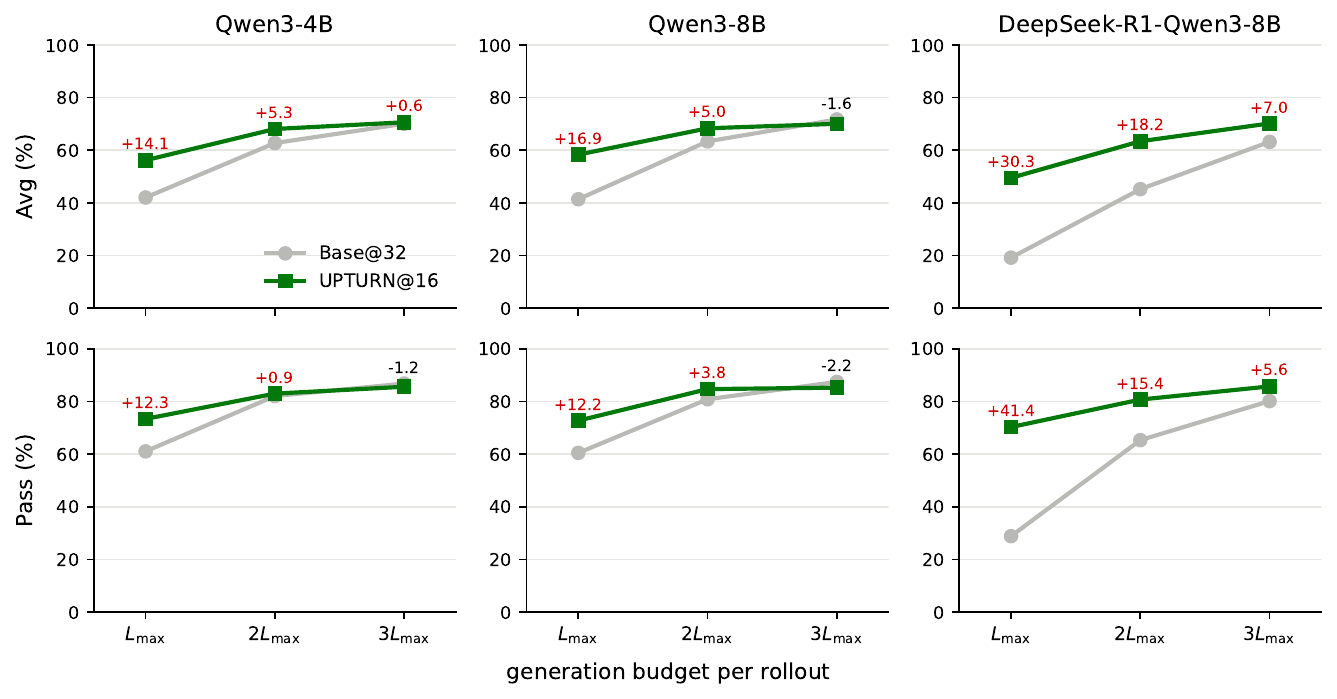}
    \vspace{-1.2em}
    \caption{\textbf{Scaling the generation budget.} Avg (top) and Pass (bottom) averaged over six benchmarks as the per-rollout budget grows from $L_{\max}$ to $3L_{\max}$; labels give UPTURN's gain over Base at each budget.}
    \label{fig:lmax-scale}
\end{figure}

UPTURN's advantage is largest at the smallest budget and narrows as the budget grows. At $L_{\max}$ it adds $14.1$ to $30.3$ Avg points across the three reasoners; at $2L_{\max}$, where Base@32 has already recovered more than $20$ Avg points, it still adds $+5.3$, $+5.0$ and $+18.2$ points on Qwen3-4B, Qwen3-8B and DeepSeek-R1-Qwen3-8B. By $3L_{\max}$ the gap has closed on the two Qwen3 reasoners, where Base@32 matches or slightly exceeds UPTURN, while DeepSeek-R1-Qwen3-8B retains $+7.0$; Base@32 extends its lead at larger budgets. These results locate UPTURN's value under tight generation budgets, where it delivers its largest improvements without allowing longer reasoning traces.

\clearpage
\section{Full Comparison with Self-Correction Baselines}
\label{app:baseline-detail}

Table~\ref{tab:baseline_detail} breaks the self-correction rows of Table~\ref{tab:budget_matched} down by reasoner and benchmark. Every row starts from the same 16 rollouts of Table~\ref{tab:main_results} and corrects each gate-flagged one once; Budget is the number of new tokens a correction may generate. Reflexion~\citep{reflexion} reflects on the first attempt in a short note and then rewrites the solution from scratch, so it is run at both $L_{\max}$ and $2L_{\max}$. SCoRe~\citep{score} conditions its second attempt on the full first attempt, so an $L_{\max}$ window leaves almost no room to write; we therefore report it at $2L_{\max}$ only. UPTURN continues from the retained prefix, and its continuation is granted the same $L_{\max}$ or $2L_{\max}$ window.

\begin{table}[h]
    \centering
    \setlength{\tabcolsep}{2pt}
    \renewcommand{\arraystretch}{1.0}
    \caption{\textbf{Full comparison with self-correction baselines}, per reasoner and benchmark, at $L_{\max}$ and $2L_{\max}$ correction budgets. $\Delta$ rows give UPTURN's gain over the best baseline at the same budget; \textbf{bold} marks the best entry per column and budget. SCoRe conditions on the full first attempt, so it is evaluated at $2L_{\max}$ only.}
    \label{tab:baseline_detail}
    \vspace{-0.5em}
    \resizebox{\textwidth}{!}{%
    \begin{tabular}{p{2.7cm}>{\footnotesize}p{1.25cm} C{0.92cm}C{0.92cm} C{0.92cm}C{0.92cm} C{0.92cm}C{0.92cm} C{0.92cm}C{0.92cm} C{0.92cm}C{0.92cm} C{0.92cm}C{0.92cm}}
        \toprule
        \multirow{2}{*}{\textbf{Method}} & \multirow{2}{*}{\normalsize\textbf{Budget}} & \multicolumn{2}{c}{\textbf{AIME24}} & \multicolumn{2}{c}{\textbf{AIME25}} & \multicolumn{2}{c}{\textbf{AIME26}} & \multicolumn{2}{c}{\textbf{AMC23}} & \multicolumn{2}{c}{\textbf{HMMT}} & \multicolumn{2}{c}{\textbf{MATH500}} \\
        \cmidrule(lr){3-4} \cmidrule(lr){5-6} \cmidrule(lr){7-8} \cmidrule(lr){9-10} \cmidrule(lr){11-12} \cmidrule(lr){13-14}
             & & {\scriptsize Avg} & {\scriptsize Pass} & {\scriptsize Avg} & {\scriptsize Pass} & {\scriptsize Avg} & {\scriptsize Pass} & {\scriptsize Avg} & {\scriptsize Pass} & {\scriptsize Avg} & {\scriptsize Pass} & {\scriptsize Avg} & {\scriptsize Pass} \\
        \midrule
        \multicolumn{14}{@{}l}{\textit{Qwen3-4B}} \\
        Reflexion@16 & $L_{\max}$ & 51.5 & 76.7 & 36.0 & 63.3 & 42.5 & 66.7 & 83.9 & 92.5 & 25.6 & 33.3 & 91.1 & \textbf{96.6} \\
        \textbf{UPTURN@16} & $L_{\max}$ & \textbf{59.0} & \textbf{83.3} & \textbf{48.3} & \textbf{70.0} & \textbf{53.8} & \textbf{76.7} & \textbf{88.4} & \textbf{97.5} & \textbf{33.9} & \textbf{45.5} & \textbf{92.1} & 96.4 \\
        \quad $\Delta$ (vs.\ best) & \cellcolor{gray!10} & \cellcolor{gray!10}\textcolor{deltacol}{+7.5} & \cellcolor{gray!10}\textcolor{deltacol}{+6.7} & \cellcolor{gray!10}\textcolor{deltacol}{+12.3} & \cellcolor{gray!10}\textcolor{deltacol}{+6.7} & \cellcolor{gray!10}\textcolor{deltacol}{+11.2} & \cellcolor{gray!10}\textcolor{deltacol}{+10.0} & \cellcolor{gray!10}\textcolor{deltacol}{+4.5} & \cellcolor{gray!10}\textcolor{deltacol}{+5.0} & \cellcolor{gray!10}\textcolor{deltacol}{+8.3} & \cellcolor{gray!10}\textcolor{deltacol}{+12.1} & \cellcolor{gray!10}\textcolor{deltacol}{+1.0} & \cellcolor{gray!10}-0.2 \\
        \cmidrule(lr){1-14}
        Reflexion@16 & $2L_{\max}$ & \textbf{64.2} & \textbf{86.7} & 49.0 & 76.7 & 55.6 & 80.0 & 90.2 & \textbf{100.0} & 34.3 & \textbf{57.6} & \textbf{93.4} & \textbf{97.8} \\
        SCoRe@16 & $2L_{\max}$ & 47.3 & \textbf{86.7} & 44.8 & \textbf{83.3} & 47.3 & \textbf{86.7} & 64.2 & \textbf{100.0} & 29.7 & 48.5 & 59.6 & 97.0 \\
        \textbf{UPTURN@16} & $2L_{\max}$ & 62.9 & \textbf{86.7} & \textbf{52.1} & 80.0 & \textbf{55.8} & 80.0 & \textbf{90.9} & 97.5 & \textbf{37.3} & \textbf{57.6} & 92.7 & 96.8 \\
        \quad $\Delta$ (vs.\ best) & \cellcolor{gray!10} & \cellcolor{gray!10}-1.3 & \cellcolor{gray!10}+0.0 & \cellcolor{gray!10}\textcolor{deltacol}{+3.1} & \cellcolor{gray!10}-3.3 & \cellcolor{gray!10}\textcolor{deltacol}{+0.2} & \cellcolor{gray!10}-6.7 & \cellcolor{gray!10}\textcolor{deltacol}{+0.8} & \cellcolor{gray!10}-2.5 & \cellcolor{gray!10}\textcolor{deltacol}{+3.0} & \cellcolor{gray!10}+0.0 & \cellcolor{gray!10}-0.7 & \cellcolor{gray!10}-1.0 \\
        \midrule\midrule
        \multicolumn{14}{@{}l}{\textit{Qwen3-8B}} \\
        Reflexion@16 & $L_{\max}$ & 39.4 & 53.3 & 24.2 & 50.0 & 27.1 & 53.3 & 67.5 & 87.5 & 20.5 & 33.3 & 84.6 & 94.6 \\
        \textbf{UPTURN@16} & $L_{\max}$ & \textbf{60.0} & \textbf{83.3} & \textbf{44.4} & \textbf{66.7} & \textbf{54.4} & \textbf{70.0} & \textbf{87.8} & \textbf{97.5} & \textbf{36.0} & \textbf{60.6} & \textbf{92.5} & \textbf{97.2} \\
        \quad $\Delta$ (vs.\ best) & \cellcolor{gray!10} & \cellcolor{gray!10}\textcolor{deltacol}{+20.6} & \cellcolor{gray!10}\textcolor{deltacol}{+30.0} & \cellcolor{gray!10}\textcolor{deltacol}{+20.2} & \cellcolor{gray!10}\textcolor{deltacol}{+16.7} & \cellcolor{gray!10}\textcolor{deltacol}{+27.3} & \cellcolor{gray!10}\textcolor{deltacol}{+16.7} & \cellcolor{gray!10}\textcolor{deltacol}{+20.3} & \cellcolor{gray!10}\textcolor{deltacol}{+10.0} & \cellcolor{gray!10}\textcolor{deltacol}{+15.5} & \cellcolor{gray!10}\textcolor{deltacol}{+27.3} & \cellcolor{gray!10}\textcolor{deltacol}{+8.0} & \cellcolor{gray!10}\textcolor{deltacol}{+2.6} \\
        \cmidrule(lr){1-14}
        Reflexion@16 & $2L_{\max}$ & 45.8 & 76.7 & 33.3 & 73.3 & 34.2 & 73.3 & 70.3 & \textbf{100.0} & 26.7 & 51.5 & 85.7 & 97.0 \\
        SCoRe@16 & $2L_{\max}$ & 63.1 & \textbf{90.0} & \textbf{50.8} & \textbf{80.0} & 54.0 & \textbf{83.3} & 72.8 & \textbf{100.0} & 36.7 & 60.6 & 67.9 & 97.4 \\
        \textbf{UPTURN@16} & $2L_{\max}$ & \textbf{63.3} & 83.3 & 49.6 & \textbf{80.0} & \textbf{56.0} & 76.7 & \textbf{88.9} & 97.5 & \textbf{38.4} & \textbf{63.6} & \textbf{93.0} & \textbf{97.6} \\
        \quad $\Delta$ (vs.\ best) & \cellcolor{gray!10} & \cellcolor{gray!10}\textcolor{deltacol}{+0.2} & \cellcolor{gray!10}-6.7 & \cellcolor{gray!10}-1.2 & \cellcolor{gray!10}+0.0 & \cellcolor{gray!10}\textcolor{deltacol}{+2.1} & \cellcolor{gray!10}-6.7 & \cellcolor{gray!10}\textcolor{deltacol}{+16.1} & \cellcolor{gray!10}-2.5 & \cellcolor{gray!10}\textcolor{deltacol}{+1.7} & \cellcolor{gray!10}\textcolor{deltacol}{+3.0} & \cellcolor{gray!10}\textcolor{deltacol}{+7.4} & \cellcolor{gray!10}\textcolor{deltacol}{+0.2} \\
        \midrule\midrule
        \multicolumn{14}{@{}l}{\textit{DeepSeek-R1-Qwen3-8B}} \\
        Reflexion@16 & $L_{\max}$ & 20.0 & 43.3 & 15.2 & 33.3 & 11.2 & 30.0 & 53.3 & 80.0 & 13.6 & 27.3 & 77.7 & 89.6 \\
        \textbf{UPTURN@16} & $L_{\max}$ & \textbf{49.4} & \textbf{76.7} & \textbf{38.5} & \textbf{66.7} & \textbf{39.8} & \textbf{56.7} & \textbf{84.1} & \textbf{97.5} & \textbf{29.5} & \textbf{51.5} & \textbf{84.8} & \textbf{96.6} \\
        \quad $\Delta$ (vs.\ best) & \cellcolor{gray!10} & \cellcolor{gray!10}\textcolor{deltacol}{+29.4} & \cellcolor{gray!10}\textcolor{deltacol}{+33.3} & \cellcolor{gray!10}\textcolor{deltacol}{+23.3} & \cellcolor{gray!10}\textcolor{deltacol}{+33.3} & \cellcolor{gray!10}\textcolor{deltacol}{+28.5} & \cellcolor{gray!10}\textcolor{deltacol}{+26.7} & \cellcolor{gray!10}\textcolor{deltacol}{+30.8} & \cellcolor{gray!10}\textcolor{deltacol}{+17.5} & \cellcolor{gray!10}\textcolor{deltacol}{+15.9} & \cellcolor{gray!10}\textcolor{deltacol}{+24.2} & \cellcolor{gray!10}\textcolor{deltacol}{+7.0} & \cellcolor{gray!10}\textcolor{deltacol}{+7.0} \\
        \cmidrule(lr){1-14}
        Reflexion@16 & $2L_{\max}$ & 45.4 & \textbf{76.7} & 36.0 & 60.0 & 36.9 & 56.7 & 84.1 & \textbf{100.0} & 23.5 & 36.4 & \textbf{87.0} & 95.4 \\
        SCoRe@16 & $2L_{\max}$ & 45.0 & 73.3 & 32.3 & 53.3 & 37.1 & 56.7 & 80.3 & 97.5 & 26.7 & 48.5 & 80.9 & 95.6 \\
        \textbf{UPTURN@16} & $2L_{\max}$ & \textbf{52.5} & \textbf{76.7} & \textbf{42.3} & \textbf{70.0} & \textbf{42.5} & \textbf{63.3} & \textbf{85.3} & 97.5 & \textbf{30.5} & \textbf{51.5} & 85.2 & \textbf{97.0} \\
        \quad $\Delta$ (vs.\ best) & \cellcolor{gray!10} & \cellcolor{gray!10}\textcolor{deltacol}{+7.1} & \cellcolor{gray!10}+0.0 & \cellcolor{gray!10}\textcolor{deltacol}{+6.2} & \cellcolor{gray!10}\textcolor{deltacol}{+10.0} & \cellcolor{gray!10}\textcolor{deltacol}{+5.4} & \cellcolor{gray!10}\textcolor{deltacol}{+6.7} & \cellcolor{gray!10}\textcolor{deltacol}{+1.2} & \cellcolor{gray!10}-2.5 & \cellcolor{gray!10}\textcolor{deltacol}{+3.8} & \cellcolor{gray!10}\textcolor{deltacol}{+3.0} & \cellcolor{gray!10}-1.8 & \cellcolor{gray!10}\textcolor{deltacol}{+1.4} \\
        \bottomrule
    \end{tabular}%
    }
\end{table}

\paragraph{Discussion.}
At $L_{\max}$, UPTURN beats Reflexion on every reasoner and benchmark in Avg, and on all but one Pass cell, by $+7.5$, $+18.6$ and $+22.5$ Avg on Qwen3-4B, Qwen3-8B and DeepSeek-R1-Qwen3-8B. Doubling the budget helps Reflexion the most, since a fresh rewrite is the method that most often runs out of room: at $2L_{\max}$ the two are within a point on Qwen3-4B, while UPTURN keeps a clear margin in Avg on Qwen3-8B and a smaller one on DeepSeek-R1-Qwen3-8B. SCoRe reaches a Pass comparable to Reflexion but a lower Avg: its rewrites are frequently truncated on the easier sets, which shows most clearly on MATH500 for Qwen3-4B and Qwen3-8B.

\clearpage
\section{Comparison with Reflection-Token Injection}
\label{app:reflect3}

Table~\ref{tab:reflect3} breaks the injection rows of Table~\ref{tab:budget_matched} down by reasoner and benchmark. AlphaOne~\citep{alphaone} and CyclicReflex~\citep{cyclicreflex} steer the same trajectory by injecting reflection tokens inline, whereas UPTURN cuts it at a learned position and re-opens a user turn; none of the three trains the reasoner.

These are not correction methods but ways of steering a single generation, so we place them in UPTURN's correction slot: UPTURN's gate selects the rollouts of the 16-rollout pool of Table~\ref{tab:main_results} that are likely wrong, and each method regenerates only those from the problem, aggregated with the same rule. The three therefore differ only in how a flagged rollout is redone. Budget is the number of new tokens a correction may generate, and every method is run at both $L_{\max}$ and $2L_{\max}$. At $L_{\max}$ the two injection methods stay close to Base@32: a regeneration that must also re-derive the discarded prefix rarely finishes within the window. With $2L_{\max}$ they improve by more than twenty points, yet UPTURN still leads at both budgets.

\begin{table}[h]
    \centering
    \setlength{\tabcolsep}{2pt}
    \renewcommand{\arraystretch}{1.0}
    \caption{\textbf{Full comparison with reflection-token injection}, per reasoner and benchmark, at $L_{\max}$ and $2L_{\max}$ correction budgets. $\Delta$ rows give UPTURN's gain over the best injection method at the same budget; \textbf{bold} marks the best entry per column and budget.}
    \label{tab:reflect3}
    \vspace{-0.5em}
    \resizebox{\textwidth}{!}{%
    \begin{tabular}{p{2.7cm}>{\footnotesize}p{1.25cm} C{0.92cm}C{0.92cm} C{0.92cm}C{0.92cm} C{0.92cm}C{0.92cm} C{0.92cm}C{0.92cm} C{0.92cm}C{0.92cm} C{0.92cm}C{0.92cm}}
        \toprule
        \multirow{2}{*}{\textbf{Method}} & \multirow{2}{*}{\normalsize\textbf{Budget}} & \multicolumn{2}{c}{\textbf{AIME24}} & \multicolumn{2}{c}{\textbf{AIME25}} & \multicolumn{2}{c}{\textbf{AIME26}} & \multicolumn{2}{c}{\textbf{AMC23}} & \multicolumn{2}{c}{\textbf{HMMT}} & \multicolumn{2}{c}{\textbf{MATH500}} \\
        \cmidrule(lr){3-4} \cmidrule(lr){5-6} \cmidrule(lr){7-8} \cmidrule(lr){9-10} \cmidrule(lr){11-12} \cmidrule(lr){13-14}
             & & {\scriptsize Avg} & {\scriptsize Pass} & {\scriptsize Avg} & {\scriptsize Pass} & {\scriptsize Avg} & {\scriptsize Pass} & {\scriptsize Avg} & {\scriptsize Pass} & {\scriptsize Avg} & {\scriptsize Pass} & {\scriptsize Avg} & {\scriptsize Pass} \\
        \midrule
        \multicolumn{14}{@{}l}{\textit{Qwen3-4B}} \\
        AlphaOne@16 & $L_{\max}$ & 33.8 & 53.3 & 21.9 & 36.7 & 27.3 & 46.7 & 67.0 & 87.5 & 16.9 & 24.2 & 84.9 & 92.4 \\
        CyclicReflex@16 & $L_{\max}$ & 37.7 & 56.7 & 25.2 & 50.0 & 28.8 & 46.7 & 70.2 & 92.5 & 19.1 & 27.3 & 86.1 & 92.6 \\
        \textbf{UPTURN@16} & $L_{\max}$ & \textbf{59.0} & \textbf{83.3} & \textbf{48.3} & \textbf{70.0} & \textbf{53.8} & \textbf{76.7} & \textbf{88.4} & \textbf{97.5} & \textbf{33.9} & \textbf{45.5} & \textbf{92.1} & \textbf{96.4} \\
        \quad $\Delta$ (vs.\ best) & \cellcolor{gray!10} & \cellcolor{gray!10}\textcolor{deltacol}{+21.2} & \cellcolor{gray!10}\textcolor{deltacol}{+26.7} & \cellcolor{gray!10}\textcolor{deltacol}{+23.1} & \cellcolor{gray!10}\textcolor{deltacol}{+20.0} & \cellcolor{gray!10}\textcolor{deltacol}{+25.0} & \cellcolor{gray!10}\textcolor{deltacol}{+30.0} & \cellcolor{gray!10}\textcolor{deltacol}{+18.3} & \cellcolor{gray!10}\textcolor{deltacol}{+5.0} & \cellcolor{gray!10}\textcolor{deltacol}{+14.8} & \cellcolor{gray!10}\textcolor{deltacol}{+18.2} & \cellcolor{gray!10}\textcolor{deltacol}{+6.1} & \cellcolor{gray!10}\textcolor{deltacol}{+3.8} \\
        \cmidrule(lr){1-14}
        AlphaOne@16 & $2L_{\max}$ & 60.6 & 83.3 & 48.3 & 63.3 & 53.3 & \textbf{80.0} & 90.5 & \textbf{100.0} & 30.3 & 48.5 & 92.3 & 96.0 \\
        CyclicReflex@16 & $2L_{\max}$ & 62.7 & 83.3 & 51.9 & 73.3 & 51.9 & 76.7 & 89.7 & \textbf{100.0} & 31.6 & 48.5 & \textbf{92.7} & 96.2 \\
        \textbf{UPTURN@16} & $2L_{\max}$ & \textbf{62.9} & \textbf{86.7} & \textbf{52.1} & \textbf{80.0} & \textbf{55.8} & \textbf{80.0} & \textbf{90.9} & 97.5 & \textbf{37.3} & \textbf{57.6} & 92.7 & \textbf{96.8} \\
        \quad $\Delta$ (vs.\ best) & \cellcolor{gray!10} & \cellcolor{gray!10}\textcolor{deltacol}{+0.2} & \cellcolor{gray!10}\textcolor{deltacol}{+3.3} & \cellcolor{gray!10}\textcolor{deltacol}{+0.2} & \cellcolor{gray!10}\textcolor{deltacol}{+6.7} & \cellcolor{gray!10}\textcolor{deltacol}{+2.5} & \cellcolor{gray!10}+0.0 & \cellcolor{gray!10}\textcolor{deltacol}{+0.5} & \cellcolor{gray!10}-2.5 & \cellcolor{gray!10}\textcolor{deltacol}{+5.7} & \cellcolor{gray!10}\textcolor{deltacol}{+9.1} & \cellcolor{gray!10}-0.1 & \cellcolor{gray!10}\textcolor{deltacol}{+0.6} \\
        \midrule\midrule
        \multicolumn{14}{@{}l}{\textit{Qwen3-8B}} \\
        AlphaOne@16 & $L_{\max}$ & 34.2 & 56.7 & 21.7 & 40.0 & 23.8 & 36.7 & 63.0 & 80.0 & 18.2 & 27.3 & 83.1 & 92.2 \\
        CyclicReflex@16 & $L_{\max}$ & 36.2 & 60.0 & 21.9 & 36.7 & 27.1 & 43.3 & 66.2 & 87.5 & 19.3 & 27.3 & 84.1 & 92.2 \\
        \textbf{UPTURN@16} & $L_{\max}$ & \textbf{60.0} & \textbf{83.3} & \textbf{44.4} & \textbf{66.7} & \textbf{54.4} & \textbf{70.0} & \textbf{87.8} & \textbf{97.5} & \textbf{36.0} & \textbf{60.6} & \textbf{92.5} & \textbf{97.2} \\
        \quad $\Delta$ (vs.\ best) & \cellcolor{gray!10} & \cellcolor{gray!10}\textcolor{deltacol}{+23.8} & \cellcolor{gray!10}\textcolor{deltacol}{+23.3} & \cellcolor{gray!10}\textcolor{deltacol}{+22.5} & \cellcolor{gray!10}\textcolor{deltacol}{+26.7} & \cellcolor{gray!10}\textcolor{deltacol}{+27.3} & \cellcolor{gray!10}\textcolor{deltacol}{+26.7} & \cellcolor{gray!10}\textcolor{deltacol}{+21.6} & \cellcolor{gray!10}\textcolor{deltacol}{+10.0} & \cellcolor{gray!10}\textcolor{deltacol}{+16.7} & \cellcolor{gray!10}\textcolor{deltacol}{+33.3} & \cellcolor{gray!10}\textcolor{deltacol}{+8.4} & \cellcolor{gray!10}\textcolor{deltacol}{+5.0} \\
        \cmidrule(lr){1-14}
        AlphaOne@16 & $2L_{\max}$ & 62.5 & \textbf{83.3} & 48.5 & 66.7 & 52.9 & \textbf{80.0} & 87.7 & \textbf{100.0} & 33.7 & 48.5 & 92.7 & 96.6 \\
        CyclicReflex@16 & $2L_{\max}$ & \textbf{63.8} & 80.0 & 48.8 & 70.0 & 55.0 & 76.7 & 87.5 & 97.5 & 34.3 & 57.6 & 93.0 & 97.0 \\
        \textbf{UPTURN@16} & $2L_{\max}$ & 63.3 & \textbf{83.3} & \textbf{49.6} & \textbf{80.0} & \textbf{56.0} & 76.7 & \textbf{88.9} & 97.5 & \textbf{38.4} & \textbf{63.6} & \textbf{93.0} & \textbf{97.6} \\
        \quad $\Delta$ (vs.\ best) & \cellcolor{gray!10} & \cellcolor{gray!10}-0.4 & \cellcolor{gray!10}+0.0 & \cellcolor{gray!10}\textcolor{deltacol}{+0.8} & \cellcolor{gray!10}\textcolor{deltacol}{+10.0} & \cellcolor{gray!10}\textcolor{deltacol}{+1.0} & \cellcolor{gray!10}-3.3 & \cellcolor{gray!10}\textcolor{deltacol}{+1.2} & \cellcolor{gray!10}-2.5 & \cellcolor{gray!10}\textcolor{deltacol}{+4.2} & \cellcolor{gray!10}\textcolor{deltacol}{+6.1} & \cellcolor{gray!10}\textcolor{deltacol}{+0.1} & \cellcolor{gray!10}\textcolor{deltacol}{+0.6} \\
        \midrule\midrule
        \multicolumn{14}{@{}l}{\textit{DeepSeek-R1-Qwen3-8B}} \\
        AlphaOne@16 & $L_{\max}$ & 4.0 & 10.0 & 3.8 & 10.0 & 2.7 & 6.7 & 26.9 & 40.0 & 0.2 & 3.0 & 53.8 & 70.6 \\
        CyclicReflex@16 & $L_{\max}$ & 7.5 & 16.7 & 9.4 & 20.0 & 4.0 & 6.7 & 32.8 & 47.5 & 1.5 & 12.1 & 65.5 & 79.0 \\
        \textbf{UPTURN@16} & $L_{\max}$ & \textbf{49.4} & \textbf{76.7} & \textbf{38.5} & \textbf{66.7} & \textbf{39.8} & \textbf{56.7} & \textbf{84.1} & \textbf{97.5} & \textbf{29.5} & \textbf{51.5} & \textbf{84.8} & \textbf{96.6} \\
        \quad $\Delta$ (vs.\ best) & \cellcolor{gray!10} & \cellcolor{gray!10}\textcolor{deltacol}{+41.9} & \cellcolor{gray!10}\textcolor{deltacol}{+60.0} & \cellcolor{gray!10}\textcolor{deltacol}{+29.2} & \cellcolor{gray!10}\textcolor{deltacol}{+46.7} & \cellcolor{gray!10}\textcolor{deltacol}{+35.8} & \cellcolor{gray!10}\textcolor{deltacol}{+50.0} & \cellcolor{gray!10}\textcolor{deltacol}{+51.2} & \cellcolor{gray!10}\textcolor{deltacol}{+50.0} & \cellcolor{gray!10}\textcolor{deltacol}{+28.0} & \cellcolor{gray!10}\textcolor{deltacol}{+39.4} & \cellcolor{gray!10}\textcolor{deltacol}{+19.2} & \cellcolor{gray!10}\textcolor{deltacol}{+17.6} \\
        \cmidrule(lr){1-14}
        AlphaOne@16 & $2L_{\max}$ & 30.4 & 53.3 & 22.3 & 40.0 & 20.2 & 46.7 & 63.6 & 87.5 & 19.3 & 27.3 & 81.6 & 91.4 \\
        CyclicReflex@16 & $2L_{\max}$ & 36.0 & 56.7 & 24.4 & 46.7 & 29.4 & 46.7 & 71.6 & 90.0 & 22.5 & 33.3 & 83.8 & 93.8 \\
        \textbf{UPTURN@16} & $2L_{\max}$ & \textbf{52.5} & \textbf{76.7} & \textbf{42.3} & \textbf{70.0} & \textbf{42.5} & \textbf{63.3} & \textbf{85.3} & \textbf{97.5} & \textbf{30.5} & \textbf{51.5} & \textbf{85.2} & \textbf{97.0} \\
        \quad $\Delta$ (vs.\ best) & \cellcolor{gray!10} & \cellcolor{gray!10}\textcolor{deltacol}{+16.5} & \cellcolor{gray!10}\textcolor{deltacol}{+20.0} & \cellcolor{gray!10}\textcolor{deltacol}{+17.9} & \cellcolor{gray!10}\textcolor{deltacol}{+23.3} & \cellcolor{gray!10}\textcolor{deltacol}{+13.1} & \cellcolor{gray!10}\textcolor{deltacol}{+16.7} & \cellcolor{gray!10}\textcolor{deltacol}{+13.8} & \cellcolor{gray!10}\textcolor{deltacol}{+7.5} & \cellcolor{gray!10}\textcolor{deltacol}{+8.0} & \cellcolor{gray!10}\textcolor{deltacol}{+18.2} & \cellcolor{gray!10}\textcolor{deltacol}{+1.4} & \cellcolor{gray!10}\textcolor{deltacol}{+3.2} \\
        \bottomrule
    \end{tabular}%
    }
\end{table}

\clearpage
\section{Complementary with Other Baselines}
\label{app:complementary}

UPTURN redirects a rollout rather than rewriting it, so it does not compete with a self-correction method for the same slot: it can run on whatever pool that method produces. Table~\ref{tab:complementary} takes the 16 corrected solutions of Reflexion~\citep{reflexion} or SCoRe~\citep{score} as the phase-1 pool, lets UPTURN's gate flag the weak ones among them, and redirects each flagged one once. Everything else, the gate, the position policy and the injected instruction, is unchanged from Table~\ref{tab:main_results}, and no component is retrained on the baseline's outputs. Reflexion is shown at both $L_{\max}$ and $2L_{\max}$, with UPTURN's continuation granted the same window as the pool it corrects. SCoRe conditions its second attempt on the full first attempt, so an $L_{\max}$ window leaves it almost no room to write and we report it at $2L_{\max}$ only.

\begin{table}[h]
    \centering
    \setlength{\tabcolsep}{2pt}
    \renewcommand{\arraystretch}{1.0}
    \caption{\textbf{UPTURN on top of a self-correction baseline}, per reasoner and benchmark. Each baseline produces its own 16 corrected solutions; UPTURN's gate then flags the weak ones among them and redirects each once, at the budget named on the baseline row. Reflexion is shown at both $L_{\max}$ and $2L_{\max}$; SCoRe conditions on the full first attempt, so only $2L_{\max}$ is meaningful for it. $\Delta$ rows give the gain of adding UPTURN; \textbf{bold} marks the better entry of the pair.}
    \label{tab:complementary}
    \vspace{-0.5em}
    \resizebox{\textwidth}{!}{%
    \begin{tabular}{p{3.6cm} C{0.92cm}C{0.92cm} C{0.92cm}C{0.92cm} C{0.92cm}C{0.92cm} C{0.92cm}C{0.92cm} C{0.92cm}C{0.92cm} C{0.92cm}C{0.92cm}}
        \toprule
        \multirow{2}{*}{\textbf{Method}} & \multicolumn{2}{c}{\textbf{AIME24}} & \multicolumn{2}{c}{\textbf{AIME25}} & \multicolumn{2}{c}{\textbf{AIME26}} & \multicolumn{2}{c}{\textbf{AMC23}} & \multicolumn{2}{c}{\textbf{HMMT}} & \multicolumn{2}{c}{\textbf{MATH500}} \\
        \cmidrule(lr){2-3} \cmidrule(lr){4-5} \cmidrule(lr){6-7} \cmidrule(lr){8-9} \cmidrule(lr){10-11} \cmidrule(lr){12-13}
             & {\scriptsize Avg} & {\scriptsize Pass} & {\scriptsize Avg} & {\scriptsize Pass} & {\scriptsize Avg} & {\scriptsize Pass} & {\scriptsize Avg} & {\scriptsize Pass} & {\scriptsize Avg} & {\scriptsize Pass} & {\scriptsize Avg} & {\scriptsize Pass} \\
        \midrule
        \multicolumn{13}{@{}l}{\textit{Qwen3-4B}} \\
        Reflexion@16 ($L_{\max}$) & \textbf{51.5} & \textbf{76.7} & \textbf{36.0} & \textbf{63.3} & \textbf{42.5} & \textbf{66.7} & 83.9 & \textbf{92.5} & \textbf{25.6} & \textbf{33.3} & \textbf{91.1} & 96.6 \\
        \quad + \textbf{UPTURN} & 49.0 & 73.3 & 35.8 & 53.3 & \textbf{42.5} & 60.0 & \textbf{84.4} & \textbf{92.5} & 23.7 & \textbf{33.3} & 90.9 & \textbf{97.0} \\
        \quad $\Delta$ & \cellcolor{gray!10}-2.5 & \cellcolor{gray!10}-3.3 & \cellcolor{gray!10}-0.2 & \cellcolor{gray!10}-10.0 & \cellcolor{gray!10}+0.0 & \cellcolor{gray!10}-6.7 & \cellcolor{gray!10}\textcolor{deltacol}{+0.5} & \cellcolor{gray!10}+0.0 & \cellcolor{gray!10}-1.9 & \cellcolor{gray!10}+0.0 & \cellcolor{gray!10}-0.3 & \cellcolor{gray!10}\textcolor{deltacol}{+0.4} \\
        \addlinespace[2pt]
        Reflexion@16 ($2L_{\max}$) & 64.2 & \textbf{86.7} & 49.0 & 76.7 & 55.6 & 80.0 & 90.2 & \textbf{100.0} & 34.3 & \textbf{57.6} & 93.4 & 97.8 \\
        \quad + \textbf{UPTURN} & \textbf{69.0} & 83.3 & \textbf{56.7} & \textbf{80.0} & \textbf{61.9} & \textbf{83.3} & \textbf{92.3} & \textbf{100.0} & \textbf{36.2} & \textbf{57.6} & \textbf{93.5} & \textbf{98.0} \\
        \quad $\Delta$ & \cellcolor{gray!10}\textcolor{deltacol}{+4.8} & \cellcolor{gray!10}-3.3 & \cellcolor{gray!10}\textcolor{deltacol}{+7.7} & \cellcolor{gray!10}\textcolor{deltacol}{+3.3} & \cellcolor{gray!10}\textcolor{deltacol}{+6.2} & \cellcolor{gray!10}\textcolor{deltacol}{+3.3} & \cellcolor{gray!10}\textcolor{deltacol}{+2.2} & \cellcolor{gray!10}+0.0 & \cellcolor{gray!10}\textcolor{deltacol}{+1.9} & \cellcolor{gray!10}+0.0 & \cellcolor{gray!10}\textcolor{deltacol}{+0.2} & \cellcolor{gray!10}\textcolor{deltacol}{+0.2} \\
        \addlinespace[2pt]
        SCoRe@16 ($2L_{\max}$) & 47.3 & 86.7 & 44.8 & \textbf{83.3} & 47.3 & \textbf{86.7} & 64.2 & \textbf{100.0} & 29.7 & \textbf{48.5} & 59.6 & \textbf{97.0} \\
        \quad + \textbf{UPTURN} & \textbf{66.9} & \textbf{90.0} & \textbf{56.7} & \textbf{83.3} & \textbf{59.8} & 83.3 & \textbf{92.2} & \textbf{100.0} & \textbf{37.3} & \textbf{48.5} & \textbf{92.5} & \textbf{97.0} \\
        \quad $\Delta$ & \cellcolor{gray!10}\textcolor{deltacol}{+19.6} & \cellcolor{gray!10}\textcolor{deltacol}{+3.3} & \cellcolor{gray!10}\textcolor{deltacol}{+11.9} & \cellcolor{gray!10}+0.0 & \cellcolor{gray!10}\textcolor{deltacol}{+12.5} & \cellcolor{gray!10}-3.3 & \cellcolor{gray!10}\textcolor{deltacol}{+28.0} & \cellcolor{gray!10}+0.0 & \cellcolor{gray!10}\textcolor{deltacol}{+7.6} & \cellcolor{gray!10}+0.0 & \cellcolor{gray!10}\textcolor{deltacol}{+32.9} & \cellcolor{gray!10}+0.0 \\
        \midrule\midrule
        \multicolumn{13}{@{}l}{\textit{Qwen3-8B}} \\
        Reflexion@16 ($L_{\max}$) & 39.4 & 53.3 & 24.2 & \textbf{50.0} & 27.1 & 53.3 & 67.5 & 87.5 & 20.5 & 33.3 & 84.6 & 94.6 \\
        \quad + \textbf{UPTURN} & \textbf{46.7} & \textbf{66.7} & \textbf{34.6} & \textbf{50.0} & \textbf{39.8} & \textbf{60.0} & \textbf{78.8} & \textbf{95.0} & \textbf{22.0} & \textbf{36.4} & \textbf{89.2} & \textbf{96.4} \\
        \quad $\Delta$ & \cellcolor{gray!10}\textcolor{deltacol}{+7.3} & \cellcolor{gray!10}\textcolor{deltacol}{+13.3} & \cellcolor{gray!10}\textcolor{deltacol}{+10.4} & \cellcolor{gray!10}+0.0 & \cellcolor{gray!10}\textcolor{deltacol}{+12.7} & \cellcolor{gray!10}\textcolor{deltacol}{+6.7} & \cellcolor{gray!10}\textcolor{deltacol}{+11.2} & \cellcolor{gray!10}\textcolor{deltacol}{+7.5} & \cellcolor{gray!10}\textcolor{deltacol}{+1.5} & \cellcolor{gray!10}\textcolor{deltacol}{+3.0} & \cellcolor{gray!10}\textcolor{deltacol}{+4.6} & \cellcolor{gray!10}\textcolor{deltacol}{+1.8} \\
        \addlinespace[2pt]
        Reflexion@16 ($2L_{\max}$) & 45.8 & 76.7 & 33.3 & 73.3 & 34.2 & 73.3 & 70.3 & \textbf{100.0} & 26.7 & 51.5 & 85.7 & 97.0 \\
        \quad + \textbf{UPTURN} & \textbf{67.7} & \textbf{90.0} & \textbf{55.8} & \textbf{76.7} & \textbf{59.2} & \textbf{80.0} & \textbf{90.6} & \textbf{100.0} & \textbf{38.6} & \textbf{57.6} & \textbf{93.2} & \textbf{97.6} \\
        \quad $\Delta$ & \cellcolor{gray!10}\textcolor{deltacol}{+21.9} & \cellcolor{gray!10}\textcolor{deltacol}{+13.3} & \cellcolor{gray!10}\textcolor{deltacol}{+22.5} & \cellcolor{gray!10}\textcolor{deltacol}{+3.3} & \cellcolor{gray!10}\textcolor{deltacol}{+25.0} & \cellcolor{gray!10}\textcolor{deltacol}{+6.7} & \cellcolor{gray!10}\textcolor{deltacol}{+20.3} & \cellcolor{gray!10}+0.0 & \cellcolor{gray!10}\textcolor{deltacol}{+11.9} & \cellcolor{gray!10}\textcolor{deltacol}{+6.1} & \cellcolor{gray!10}\textcolor{deltacol}{+7.5} & \cellcolor{gray!10}\textcolor{deltacol}{+0.6} \\
        \addlinespace[2pt]
        SCoRe@16 ($2L_{\max}$) & 63.1 & \textbf{90.0} & 50.8 & \textbf{80.0} & 54.0 & \textbf{83.3} & 72.8 & \textbf{100.0} & 36.7 & 60.6 & 67.9 & 97.4 \\
        \quad + \textbf{UPTURN} & \textbf{71.7} & \textbf{90.0} & \textbf{55.8} & 76.7 & \textbf{58.8} & \textbf{83.3} & \textbf{91.7} & \textbf{100.0} & \textbf{41.3} & \textbf{63.6} & \textbf{92.2} & \textbf{97.6} \\
        \quad $\Delta$ & \cellcolor{gray!10}\textcolor{deltacol}{+8.5} & \cellcolor{gray!10}+0.0 & \cellcolor{gray!10}\textcolor{deltacol}{+5.0} & \cellcolor{gray!10}-3.3 & \cellcolor{gray!10}\textcolor{deltacol}{+4.8} & \cellcolor{gray!10}+0.0 & \cellcolor{gray!10}\textcolor{deltacol}{+18.9} & \cellcolor{gray!10}+0.0 & \cellcolor{gray!10}\textcolor{deltacol}{+4.5} & \cellcolor{gray!10}\textcolor{deltacol}{+3.0} & \cellcolor{gray!10}\textcolor{deltacol}{+24.3} & \cellcolor{gray!10}\textcolor{deltacol}{+0.2} \\
        \midrule\midrule
        \multicolumn{13}{@{}l}{\textit{DeepSeek-R1-Qwen3-8B}} \\
        Reflexion@16 ($L_{\max}$) & 20.0 & 43.3 & 15.2 & 33.3 & 11.2 & 30.0 & 53.3 & 80.0 & 13.6 & 27.3 & 77.7 & 89.6 \\
        \quad + \textbf{UPTURN} & \textbf{39.0} & \textbf{66.7} & \textbf{29.8} & \textbf{46.7} & \textbf{31.9} & \textbf{50.0} & \textbf{69.5} & \textbf{95.0} & \textbf{18.8} & \textbf{39.4} & \textbf{78.3} & \textbf{94.2} \\
        \quad $\Delta$ & \cellcolor{gray!10}\textcolor{deltacol}{+19.0} & \cellcolor{gray!10}\textcolor{deltacol}{+23.3} & \cellcolor{gray!10}\textcolor{deltacol}{+14.6} & \cellcolor{gray!10}\textcolor{deltacol}{+13.3} & \cellcolor{gray!10}\textcolor{deltacol}{+20.6} & \cellcolor{gray!10}\textcolor{deltacol}{+20.0} & \cellcolor{gray!10}\textcolor{deltacol}{+16.2} & \cellcolor{gray!10}\textcolor{deltacol}{+15.0} & \cellcolor{gray!10}\textcolor{deltacol}{+5.1} & \cellcolor{gray!10}\textcolor{deltacol}{+12.1} & \cellcolor{gray!10}\textcolor{deltacol}{+0.6} & \cellcolor{gray!10}\textcolor{deltacol}{+4.6} \\
        \addlinespace[2pt]
        Reflexion@16 ($2L_{\max}$) & 45.4 & 76.7 & 36.0 & 60.0 & 36.9 & 56.7 & \textbf{84.1} & \textbf{100.0} & 23.5 & 36.4 & \textbf{87.0} & 95.4 \\
        \quad + \textbf{UPTURN} & \textbf{54.0} & \textbf{80.0} & \textbf{45.8} & \textbf{63.3} & \textbf{45.6} & \textbf{63.3} & 81.9 & \textbf{100.0} & \textbf{29.9} & \textbf{42.4} & 82.0 & \textbf{96.6} \\
        \quad $\Delta$ & \cellcolor{gray!10}\textcolor{deltacol}{+8.5} & \cellcolor{gray!10}\textcolor{deltacol}{+3.3} & \cellcolor{gray!10}\textcolor{deltacol}{+9.8} & \cellcolor{gray!10}\textcolor{deltacol}{+3.3} & \cellcolor{gray!10}\textcolor{deltacol}{+8.8} & \cellcolor{gray!10}\textcolor{deltacol}{+6.7} & \cellcolor{gray!10}-2.2 & \cellcolor{gray!10}+0.0 & \cellcolor{gray!10}\textcolor{deltacol}{+6.4} & \cellcolor{gray!10}\textcolor{deltacol}{+6.1} & \cellcolor{gray!10}-5.0 & \cellcolor{gray!10}\textcolor{deltacol}{+1.2} \\
        \addlinespace[2pt]
        SCoRe@16 ($2L_{\max}$) & 45.0 & 73.3 & 32.3 & 53.3 & 37.1 & 56.7 & 80.3 & 97.5 & 26.7 & \textbf{48.5} & 80.9 & 95.6 \\
        \quad + \textbf{UPTURN} & \textbf{52.9} & \textbf{83.3} & \textbf{40.4} & \textbf{63.3} & \textbf{44.2} & \textbf{66.7} & \textbf{84.4} & \textbf{100.0} & \textbf{29.7} & \textbf{48.5} & \textbf{84.6} & \textbf{96.4} \\
        \quad $\Delta$ & \cellcolor{gray!10}\textcolor{deltacol}{+7.9} & \cellcolor{gray!10}\textcolor{deltacol}{+10.0} & \cellcolor{gray!10}\textcolor{deltacol}{+8.1} & \cellcolor{gray!10}\textcolor{deltacol}{+10.0} & \cellcolor{gray!10}\textcolor{deltacol}{+7.1} & \cellcolor{gray!10}\textcolor{deltacol}{+10.0} & \cellcolor{gray!10}\textcolor{deltacol}{+4.1} & \cellcolor{gray!10}\textcolor{deltacol}{+2.5} & \cellcolor{gray!10}\textcolor{deltacol}{+3.0} & \cellcolor{gray!10}+0.0 & \cellcolor{gray!10}\textcolor{deltacol}{+3.7} & \cellcolor{gray!10}\textcolor{deltacol}{+0.8} \\
        \bottomrule
    \end{tabular}%
    }
\end{table}

Adding UPTURN improves every baseline on every reasoner, averaged over benchmarks: $+6.6$ Avg and $+5.6$ Pass on Reflexion at $L_{\max}$, $+8.8$ and $+3.0$ on Reflexion at $2L_{\max}$, and $+11.8$ and $+1.8$ on SCoRe. The gain is largest where the pool it corrects is weakest. On Reflexion at $L_{\max}$, where a fresh rewrite most often runs out of room, it adds $+12.7$ Avg on DeepSeek-R1-Qwen3-8B against $+3.6$ on the two Qwen3 reasoners; doubling the budget reverses the ordering, to $+4.4$ on DeepSeek-R1-Qwen3-8B against $+11.0$ on Qwen3. Where the pool is already strong the gain narrows and can turn slightly negative, as on AMC23 and MATH500 for DeepSeek-R1-Qwen3-8B at $2L_{\max}$, on which Reflexion scores $84.1$ and $87.0$: a cut placed inside an already-correct rewrite can only lose. Redirecting a rollout therefore composes with rewriting it, and pays off most where the rewrite had the least room to finish.

\clearpage
\section{Seed Robustness}
\label{app:seedrep}

Table~\ref{tab:seedrep} repeats Table~\ref{tab:main_results} with three seeds, each used both to train the position policy and to draw the rollouts. The standard deviation of Avg stays within about two points, far below every gain of UPTURN over Base.

\begin{table}[h]
    \centering
    \setlength{\tabcolsep}{3pt}
    \renewcommand{\arraystretch}{0.85}
    \footnotesize
    \caption{\textbf{Seed robustness.} Table~\ref{tab:main_results} repeated with three seeds for training the position policy and drawing rollouts; mean\,$\pm$\,std over the three runs. $\Delta$ is UPTURN's gain; \textbf{bold} marks the higher mean.}
    \label{tab:seedrep}
    \vspace{-0.5em}
    \begin{tabular}{p{2.4cm} C{1.45cm}C{1.45cm} C{3pt} C{1.45cm}C{1.45cm} C{3pt} C{1.45cm}C{1.45cm}}
        \toprule
        \multirow{2}{*}{\textbf{Method}} & \multicolumn{2}{c}{\textbf{AIME24}} & & \multicolumn{2}{c}{\textbf{AIME25}} & & \multicolumn{2}{c}{\textbf{AIME26}} \\
        \cmidrule(lr){2-3} \cmidrule(lr){5-6} \cmidrule(lr){8-9}
             & {\scriptsize Avg} & {\scriptsize Pass} & & {\scriptsize Avg} & {\scriptsize Pass} & & {\scriptsize Avg} & {\scriptsize Pass} \\
        \midrule
        \multicolumn{8}{@{}l}{\textit{Qwen3-4B}} \\
        \quad Base@32 & 33.4{\scriptsize$\pm$0.8} & 62.2{\scriptsize$\pm$1.9} & & 21.6{\scriptsize$\pm$0.2} & 47.8{\scriptsize$\pm$1.9} & & 26.1{\scriptsize$\pm$1.2} & 43.3{\scriptsize$\pm$0.0} \\
        \quad \textbf{UPTURN@16} & \textbf{52.1{\scriptsize$\pm$1.1}} & \textbf{81.1{\scriptsize$\pm$3.8}} & & \textbf{40.0{\scriptsize$\pm$1.0}} & \textbf{58.9{\scriptsize$\pm$1.9}} & & \textbf{45.8{\scriptsize$\pm$1.5}} & \textbf{70.0{\scriptsize$\pm$0.0}} \\
        \quad $\Delta$ & \cellcolor{gray!10}\textcolor{deltacol}{+18.7} & \cellcolor{gray!10}\textcolor{deltacol}{+18.9} & \cellcolor{gray!10} & \cellcolor{gray!10}\textcolor{deltacol}{+18.4} & \cellcolor{gray!10}\textcolor{deltacol}{+11.1} & \cellcolor{gray!10} & \cellcolor{gray!10}\textcolor{deltacol}{+19.7} & \cellcolor{gray!10}\textcolor{deltacol}{+26.7} \\
        \addlinespace[1pt]
        \multicolumn{8}{@{}l}{\textit{Qwen3-8B}} \\
        \quad Base@32 & 35.0{\scriptsize$\pm$1.0} & 61.1{\scriptsize$\pm$5.1} & & 21.3{\scriptsize$\pm$0.6} & 43.3{\scriptsize$\pm$3.3} & & 25.3{\scriptsize$\pm$0.6} & 46.7{\scriptsize$\pm$3.3} \\
        \quad \textbf{UPTURN@16} & \textbf{53.2{\scriptsize$\pm$0.6}} & \textbf{78.9{\scriptsize$\pm$7.7}} & & \textbf{39.2{\scriptsize$\pm$0.9}} & \textbf{55.6{\scriptsize$\pm$1.9}} & & \textbf{46.6{\scriptsize$\pm$1.3}} & \textbf{68.9{\scriptsize$\pm$1.9}} \\
        \quad $\Delta$ & \cellcolor{gray!10}\textcolor{deltacol}{+18.2} & \cellcolor{gray!10}\textcolor{deltacol}{+17.8} & \cellcolor{gray!10} & \cellcolor{gray!10}\textcolor{deltacol}{+17.9} & \cellcolor{gray!10}\textcolor{deltacol}{+12.2} & \cellcolor{gray!10} & \cellcolor{gray!10}\textcolor{deltacol}{+21.3} & \cellcolor{gray!10}\textcolor{deltacol}{+22.2} \\
        \addlinespace[1pt]
        \multicolumn{8}{@{}l}{\textit{DeepSeek-R1-Qwen3-8B}} \\
        \quad Base@32 & 6.1{\scriptsize$\pm$0.4} & 17.8{\scriptsize$\pm$1.9} & & 6.6{\scriptsize$\pm$0.5} & 14.4{\scriptsize$\pm$1.9} & & 3.4{\scriptsize$\pm$0.1} & 6.7{\scriptsize$\pm$0.0} \\
        \quad \textbf{UPTURN@16} & \textbf{41.5{\scriptsize$\pm$2.0}} & \textbf{64.4{\scriptsize$\pm$6.9}} & & \textbf{31.5{\scriptsize$\pm$0.3}} & \textbf{53.3{\scriptsize$\pm$0.0}} & & \textbf{33.4{\scriptsize$\pm$1.9}} & \textbf{57.8{\scriptsize$\pm$1.9}} \\
        \quad $\Delta$ & \cellcolor{gray!10}\textcolor{deltacol}{+35.3} & \cellcolor{gray!10}\textcolor{deltacol}{+46.7} & \cellcolor{gray!10} & \cellcolor{gray!10}\textcolor{deltacol}{+25.0} & \cellcolor{gray!10}\textcolor{deltacol}{+38.9} & \cellcolor{gray!10} & \cellcolor{gray!10}\textcolor{deltacol}{+30.0} & \cellcolor{gray!10}\textcolor{deltacol}{+51.1} \\
        \midrule\midrule
        \multirow{2}{*}{\textbf{Method}} & \multicolumn{2}{c}{\textbf{AMC23}} & & \multicolumn{2}{c}{\textbf{HMMT}} & & \multicolumn{2}{c}{\textbf{MATH500}} \\
        \cmidrule(lr){2-3} \cmidrule(lr){5-6} \cmidrule(lr){8-9}
             & {\scriptsize Avg} & {\scriptsize Pass} & & {\scriptsize Avg} & {\scriptsize Pass} & & {\scriptsize Avg} & {\scriptsize Pass} \\
        \midrule
        \multicolumn{8}{@{}l}{\textit{Qwen3-4B}} \\
        \quad Base@32 & 68.3{\scriptsize$\pm$1.2} & 92.5{\scriptsize$\pm$0.0} & & 19.5{\scriptsize$\pm$0.5} & 26.3{\scriptsize$\pm$1.7} & & 84.9{\scriptsize$\pm$0.1} & 93.5{\scriptsize$\pm$0.3} \\
        \quad \textbf{UPTURN@16} & \textbf{84.9{\scriptsize$\pm$1.3}} & \textbf{95.8{\scriptsize$\pm$1.4}} & & \textbf{27.1{\scriptsize$\pm$1.3}} & \textbf{39.4{\scriptsize$\pm$3.0}} & & \textbf{91.0{\scriptsize$\pm$0.2}} & \textbf{96.0{\scriptsize$\pm$0.3}} \\
        \quad $\Delta$ & \cellcolor{gray!10}\textcolor{deltacol}{+16.7} & \cellcolor{gray!10}\textcolor{deltacol}{+3.3} & \cellcolor{gray!10} & \cellcolor{gray!10}\textcolor{deltacol}{+7.6} & \cellcolor{gray!10}\textcolor{deltacol}{+13.1} & \cellcolor{gray!10} & \cellcolor{gray!10}\textcolor{deltacol}{+6.1} & \cellcolor{gray!10}\textcolor{deltacol}{+2.5} \\
        \addlinespace[1pt]
        \multicolumn{8}{@{}l}{\textit{Qwen3-8B}} \\
        \quad Base@32 & 64.6{\scriptsize$\pm$0.4} & 89.2{\scriptsize$\pm$2.9} & & 18.9{\scriptsize$\pm$0.3} & 27.3{\scriptsize$\pm$0.0} & & 83.2{\scriptsize$\pm$0.1} & 93.1{\scriptsize$\pm$0.4} \\
        \quad \textbf{UPTURN@16} & \textbf{84.5{\scriptsize$\pm$0.8}} & \textbf{95.8{\scriptsize$\pm$1.4}} & & \textbf{31.6{\scriptsize$\pm$0.5}} & \textbf{47.5{\scriptsize$\pm$4.6}} & & \textbf{91.8{\scriptsize$\pm$0.1}} & \textbf{96.5{\scriptsize$\pm$0.1}} \\
        \quad $\Delta$ & \cellcolor{gray!10}\textcolor{deltacol}{+20.0} & \cellcolor{gray!10}\textcolor{deltacol}{+6.7} & \cellcolor{gray!10} & \cellcolor{gray!10}\textcolor{deltacol}{+12.7} & \cellcolor{gray!10}\textcolor{deltacol}{+20.2} & \cellcolor{gray!10} & \cellcolor{gray!10}\textcolor{deltacol}{+8.7} & \cellcolor{gray!10}\textcolor{deltacol}{+3.4} \\
        \addlinespace[1pt]
        \multicolumn{8}{@{}l}{\textit{DeepSeek-R1-Qwen3-8B}} \\
        \quad Base@32 & 32.2{\scriptsize$\pm$0.3} & 49.2{\scriptsize$\pm$2.9} & & 2.0{\scriptsize$\pm$0.2} & 13.1{\scriptsize$\pm$3.5} & & 64.5{\scriptsize$\pm$0.0} & 80.6{\scriptsize$\pm$0.8} \\
        \quad \textbf{UPTURN@16} & \textbf{78.6{\scriptsize$\pm$0.9}} & \textbf{96.7{\scriptsize$\pm$1.4}} & & \textbf{25.8{\scriptsize$\pm$1.1}} & \textbf{44.4{\scriptsize$\pm$4.6}} & & \textbf{82.7{\scriptsize$\pm$0.5}} & \textbf{95.3{\scriptsize$\pm$0.2}} \\
        \quad $\Delta$ & \cellcolor{gray!10}\textcolor{deltacol}{+46.4} & \cellcolor{gray!10}\textcolor{deltacol}{+47.5} & \cellcolor{gray!10} & \cellcolor{gray!10}\textcolor{deltacol}{+23.9} & \cellcolor{gray!10}\textcolor{deltacol}{+31.3} & \cellcolor{gray!10} & \cellcolor{gray!10}\textcolor{deltacol}{+18.2} & \cellcolor{gray!10}\textcolor{deltacol}{+14.7} \\
        \bottomrule
    \end{tabular}
\end{table}

\clearpage
\section{Results on Other Models}
\label{app:other-models}

Table~\ref{tab:other_models} reports results on two additional RL-finetuned thinking reasoners under the identical setup as Table~\ref{tab:main_results}: AceReason-Nemotron-1.1-7B and Skywork-OR1-7B. AceReason follows the Qwen ChatML template and Skywork the DeepSeek template, so the turn boundary is injected with the corresponding sequences of Table~\ref{tab:mt_format}. On both, UPTURN improves average accuracy on every benchmark ($+1.8$ to $+15.2$ points) and pass rate on all but two benchmarks, where it ties Base.

\begin{table}[h]
    \centering
    \setlength{\tabcolsep}{2pt}
    \renewcommand{\arraystretch}{1.0}
    \caption{\textbf{Results on two additional reasoners}, with the same setup as Table~\ref{tab:main_results}. $\Delta$ denotes UPTURN's gain over Base; \textbf{bold} marks improvements.}
    \vspace{-0.5em}
    \resizebox{\textwidth}{!}{%
    \begin{tabular}{p{3.3cm}
        C{0.9cm}C{0.9cm} C{3pt}
        C{0.9cm}C{0.9cm} C{3pt}
        C{0.9cm}C{0.9cm} C{3pt}
        C{0.9cm}C{0.9cm} C{3pt}
        C{0.9cm}C{0.9cm} C{3pt}
        C{0.9cm}C{0.9cm}}
        \toprule
        \multirow{2}{*}{\textbf{Method}}
            & \multicolumn{2}{c}{\textbf{AIME24}} &
            & \multicolumn{2}{c}{\textbf{AIME25}} &
            & \multicolumn{2}{c}{\textbf{AIME26}} &
            & \multicolumn{2}{c}{\textbf{AMC23}} &
            & \multicolumn{2}{c}{\textbf{HMMT}} &
            & \multicolumn{2}{c}{\textbf{MATH500}} \\
        \cmidrule(lr){2-3} \cmidrule(lr){5-6} \cmidrule(lr){8-9} \cmidrule(lr){11-12} \cmidrule(lr){14-15} \cmidrule(lr){17-18}
             & {\scriptsize Avg} & {\scriptsize Pass} &
            & {\scriptsize Avg} & {\scriptsize Pass} &
            & {\scriptsize Avg} & {\scriptsize Pass} &
            & {\scriptsize Avg} & {\scriptsize Pass} &
            & {\scriptsize Avg} & {\scriptsize Pass} &
            & {\scriptsize Avg} & {\scriptsize Pass} \\
        \midrule

        \multicolumn{18}{@{}l}{\textit{AceReason-Nemotron-1.1-7B}} \\
        \quad Base@32 & 37.2 & 73.3 & & 27.7 & 40.0 & & 31.5 & 56.7 & & 79.7 & 95.0 & & 19.5 & 30.3 & & 88.4 & 96.2 \\
        \quad \textbf{UPTURN@16} & \textbf{47.5} & 73.3 & & \textbf{32.5} & \textbf{46.7} & & \textbf{44.6} & \textbf{63.3} & & \textbf{89.7} & 95.0 & & \textbf{27.1} & \textbf{45.5} & & \textbf{92.1} & \textbf{97.0} \\
        \quad $\Delta$ & \cellcolor{gray!10}\textcolor{deltacol}{+10.3} & \cellcolor{gray!10}{0.0} & \cellcolor{gray!10} & \cellcolor{gray!10}\textcolor{deltacol}{+4.8} & \cellcolor{gray!10}\textcolor{deltacol}{+6.7} & \cellcolor{gray!10} & \cellcolor{gray!10}\textcolor{deltacol}{+13.1} & \cellcolor{gray!10}\textcolor{deltacol}{+6.7} & \cellcolor{gray!10} & \cellcolor{gray!10}\textcolor{deltacol}{+10.0} & \cellcolor{gray!10}{0.0} & \cellcolor{gray!10} & \cellcolor{gray!10}\textcolor{deltacol}{+7.6} & \cellcolor{gray!10}\textcolor{deltacol}{+15.2} & \cellcolor{gray!10} & \cellcolor{gray!10}\textcolor{deltacol}{+3.7} & \cellcolor{gray!10}\textcolor{deltacol}{+0.8} \\
        \addlinespace[3pt]
        \midrule

        \multicolumn{18}{@{}l}{\textit{Skywork-OR1-7B}} \\
        \quad Base@32 & 33.8 & 70.0 & & 23.3 & 40.0 & & 27.1 & 50.0 & & 67.6 & 90.0 & & 19.5 & 33.3 & & 83.9 & 94.6 \\
        \quad \textbf{UPTURN@16} & \textbf{49.0} & \textbf{73.3} & & \textbf{34.2} & \textbf{56.7} & & \textbf{40.0} & \textbf{66.7} & & \textbf{81.1} & \textbf{95.0} & & \textbf{27.3} & \textbf{42.4} & & \textbf{85.7} & \textbf{97.6} \\
        \quad $\Delta$ & \cellcolor{gray!10}\textcolor{deltacol}{+15.2} & \cellcolor{gray!10}\textcolor{deltacol}{+3.3} & \cellcolor{gray!10} & \cellcolor{gray!10}\textcolor{deltacol}{+10.8} & \cellcolor{gray!10}\textcolor{deltacol}{+16.7} & \cellcolor{gray!10} & \cellcolor{gray!10}\textcolor{deltacol}{+12.9} & \cellcolor{gray!10}\textcolor{deltacol}{+16.7} & \cellcolor{gray!10} & \cellcolor{gray!10}\textcolor{deltacol}{+13.5} & \cellcolor{gray!10}\textcolor{deltacol}{+5.0} & \cellcolor{gray!10} & \cellcolor{gray!10}\textcolor{deltacol}{+7.8} & \cellcolor{gray!10}\textcolor{deltacol}{+9.1} & \cellcolor{gray!10} & \cellcolor{gray!10}\textcolor{deltacol}{+1.8} & \cellcolor{gray!10}\textcolor{deltacol}{+3.0} \\
        \bottomrule
    \end{tabular}%
    }
    \label{tab:other_models}
\end{table}

\clearpage
\includepdf[pages=1, scale=0.8, offset=0 -1.5cm, pagecommand={\section{Full Rollouts for Figure~\ref{fig:motivation}}\label{app:full-rollouts}\thispagestyle{plain}}]{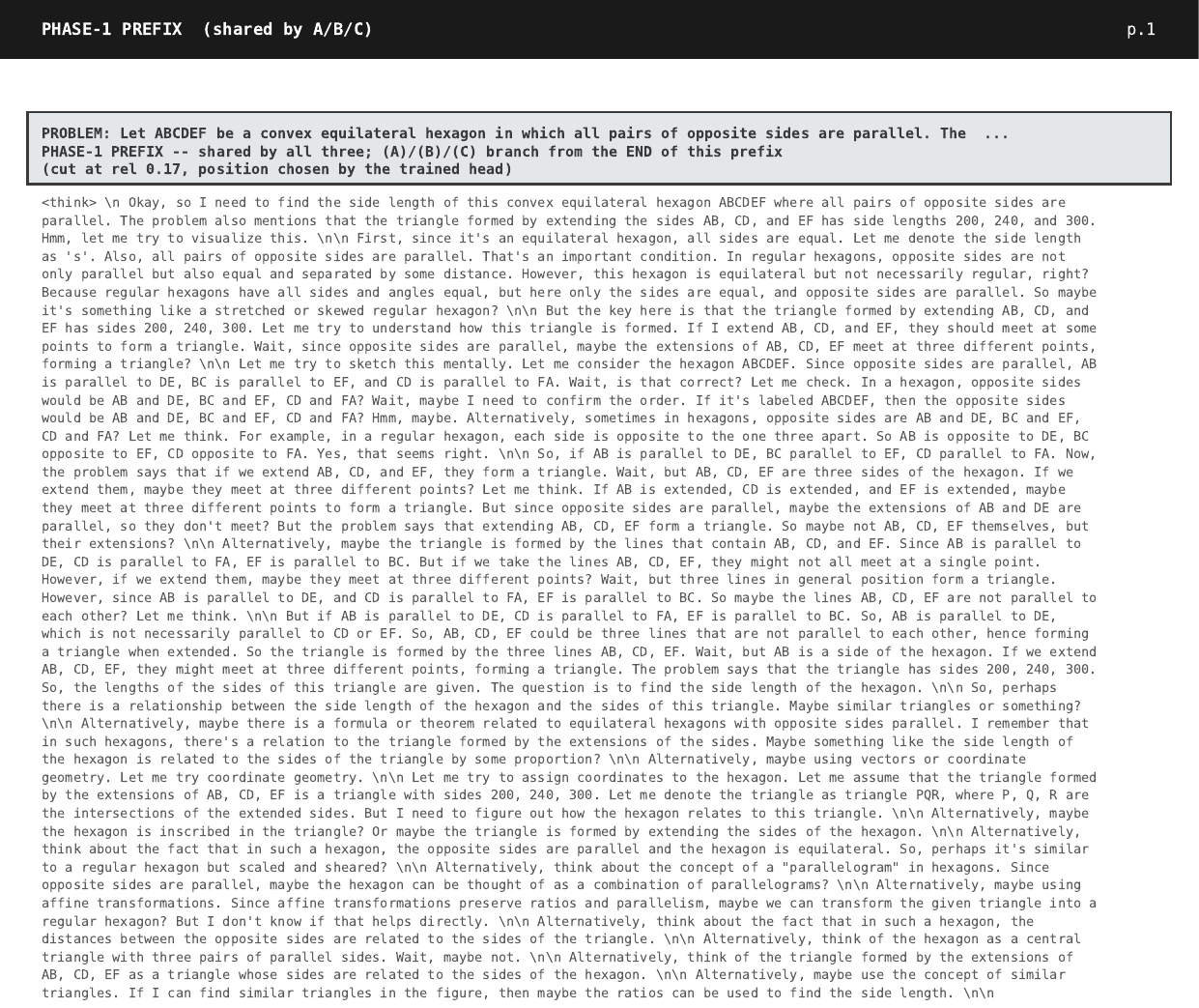}
\includepdf[pages=2-, scale=0.8, pagecommand={\thispagestyle{plain}}]{figure/appendix_full_rollouts.pdf}

\end{document}